\documentclass{article} 
\usepackage{preprint,times}

\usepackage{amsmath,amsfonts,bm}

\def\eqref#1{equation~\ref{#1}}

\def\1{\bm{1}}

\DeclareMathAlphabet{\mathsfit}{\encodingdefault}{\sfdefault}{m}{sl}
\SetMathAlphabet{\mathsfit}{bold}{\encodingdefault}{\sfdefault}{bx}{n}

\usepackage{hyperref}
\usepackage{url}

\usepackage{booktabs}       
\usepackage{amsfonts}       
\usepackage{nicefrac}       
\usepackage{microtype}      
\usepackage{lipsum}         
\usepackage{graphicx}
\usepackage{natbib}
\usepackage{doi}

\usepackage{algorithm}
\usepackage[noend]{algpseudocode}
\usepackage{amsmath}
\usepackage{amssymb}
\usepackage{amsthm}
\usepackage{calligra}
\usepackage{caption}
\usepackage{cleveref}       
\usepackage{enumitem}
\usepackage[T1]{fontenc}
\usepackage{graphicx}
\usepackage{makecell}
\usepackage{mathtools}
\usepackage{multirow}
\usepackage{wrapfig}
\usepackage{siunitx}
\usepackage{subcaption}
\usepackage{pifont}
\usepackage{utfsym}
\usepackage[table]{xcolor}
\usepackage{fontawesome5}

\newtheorem{theorem}{Theorem}[]

\newtheorem{definition}[theorem]{Definition}

\theoremstyle{remark}

\definecolor{ipsdgray}{gray}{0.94}

\newcommand{\xheavyseven}{%
  \cmidrule[\heavyrulewidth](l{-\tabcolsep}r{-\tabcolsep}){1-7}}
\newcommand{\xlightseven}{%
  \cmidrule[\lightrulewidth](l{-\tabcolsep}r{-\tabcolsep}){1-7}}

\newcommand{\chg}[1]{\,{\scriptsize\textcolor{gray}{(#1)}}}
\newcommand{\hlei}{\textsc{Prim}}
\newcommand{\abs}{\textsc{Absorb}}

\definecolor{avgcol}{HTML}{1A5276}

\usepackage[table]{xcolor}  
\definecolor{myred}{HTML}{E14169} 
\definecolor{royalblue}{HTML}{4169E1}
\definecolor{mycyan}{HTML}{0AB6D8}
\definecolor{myblue}{HTML}{2766F5}  
\definecolor{mydeepblue}{HTML}{004E98} 
\definecolor{mygreen}{HTML}{D1FFBD}

\usepackage[utf8]{inputenc} 
\usepackage[T1]{fontenc}    
\usepackage{booktabs}       
\usepackage{amsfonts}       
\usepackage{nicefrac}       
\usepackage{microtype}      
\usepackage{proof-at-the-end}  
\usepackage{multirow}
\usepackage{lscape}         

\usepackage{graphicx,wrapfig}
\usepackage{amsfonts}
\usepackage{url}
\usepackage{enumitem}
\usepackage{amsthm}
\usepackage{thmtools}
\usepackage{thm-restate}

\usepackage{listings}
\definecolor{codecomment}{rgb}{0,0.6,0}   
\definecolor{codekeyword}{rgb}{0,0,1}     
\definecolor{codestring}{rgb}{0.8,0,0}    

\lstdefinestyle{mypython}{
    language=Python,
    backgroundcolor=\color{royalblue!5!white},   
    basicstyle=\ttfamily\tiny, 
    keywordstyle=\color{codekeyword}\bfseries,
    stringstyle=\color{codestring},
    commentstyle=\color{codecomment}\itshape,
    numbers=left,
    numberstyle=\tiny\color{gray},
    frame=single,                     
    rulecolor=\color{black},          
    breaklines=true,                  
    showstringspaces=false
}

\usepackage[many]{tcolorbox}
\newtcolorbox{boxK}[2][]{
    sharpish corners, 
    boxrule = 0pt,
    toprule = 4.5pt, 
    enhanced,
    fuzzy shadow = {0pt}{-2pt}{-0.5pt}{0.5pt}{black!35}, 
    fontupper = \ttfamily\small, 
    boxsep = 5pt, 
    left = 5pt, 
    right = 5pt, 
    top = 5pt, 
    bottom = 5pt, 
    #1                       
}

\newtcolorbox{takeawaybox}[2][]{
    enhanced,
    boxsep = 2pt, 
    left = 2pt, 
    right = 2pt, 
    top = 2pt, 
    bottom = 2pt, 
    #1                       
} 

\newcommand{\bc}{\begin{center}}
\newcommand{\ec}{\end{center}}

\newcommand{\bdm}{\begin{displaymath}}
\newcommand{\edm}{\end{displaymath}}

\newcommand{\beq}{\begin{equation}}
\newcommand{\eeq}{\end{equation}}

\newcommand{\bfl}{\begin{flushleft}}
\newcommand{\efl}{\end{flushleft}}

\newcommand{\bt}{\begin{tabbing}}
\newcommand{\et}{\end{tabbing}}

\newcommand{\beqn}{\begin{align}}
\newcommand{\eeqn}{\end{align}}

\newcommand{\beqs}{\begin{align*}} 
\newcommand{\eeqs}{\end{align*}}  

\iclrfinalcopy
\title{The Missing Primitive: Diagnosing and Repairing Mathematical Reasoning in Large Language Models}

\author{
Shuo Xing$^1$\quad 
Zilin Dai$^2$\quad 
Chengyuan Qian$^1$\quad
Fangzhou Lin$^1$\quad 
Wenjing Chen$^1$\quad \\
\bf Ping He$^3$\quad 
Pan Lu$^4$\quad
Alvaro Velasquez$^5$\quad
Mohit Bansal$^6$\quad
Zhengzhong Tu$^1$\thanks{\  Corresponding author.}\\
\\
$^1$Texas A\&M University \quad
$^2$Harvard University \quad
$^3$Vanderbilt University \quad
\\
$^4$Stanford University \quad
$^5$DARPA \quad
$^6$University of North Carolina at Chapel Hill \quad
\\
\\
\texttt{\{shuoxing,tzz\}@tamu.edu}
\\
\\
\faGlobe~Model \& Code: \url{https://taco-group.github.io/Math-Primitive/}
\vspace{-2em}
}

\begin{document}

\maketitle

\begin{abstract}
While Large Language Models (LLMs) have demonstrated striking capabilities on frontier mathematical problems, it remains unclear whether they possess the structural mathematical understanding underlying their solutions. In this paper, we take a first step toward systematically studying mathematical understanding in LLMs, from diagnosing its distinct capabilities to leveraging these findings to improve post-training. First, we introduce the notion of Mathematical Primitive to probe structural mathematical understanding and propose \hlei{}, a novel benchmark that evaluates mathematical reasoning along four distinct dimensions: Discovery, Generation, Digestion, and Execution. 
Second, our systematic diagnosis shows that solution accuracy masks distinct capability profiles, primitives unlock substantial latent execution capacity, and Discovery is the dominant bottleneck in mathematical reasoning.
Our post-training analysis further shows that discovery-limited failures are particularly amenable to repair. Finally, building on these findings, we introduce \abs{}, a primitive-privileged self-distillation framework that selectively transfers primitive-guided reasoning into the student model. Extensive experiments demonstrate that \abs{} consistently improves mathematical reasoning over baselines across model scales and challenging benchmarks. 
\end{abstract}

\section{Introduction}

Recent advances in Large Language Models (LLMs) have driven rapid progress in mathematical problem solving, with frontier models increasingly capable of tackling research-level questions and, in several recent cases, resolving or making substantive progress on long-standing open problems~\citep{openai2026tenadvances,openai2026unitdistance,claude2026riemann,bubeck2025early,jang2025point}. Yet these striking successes leave unresolved a more foundational question: what models actually understand about the problems they solve and the mathematical content of the solutions they produce. Successful problem solving does not necessarily imply such understanding: models may arrive at correct answers through specious deductions~\citep{boye2025large}, while conversely identifying sound high-level proof structures yet failing in low-level technical execution~\citep{lu2024proof,qin2025decomposing}. Existing works provide limited insight into this disconnect, as mathematical reasoning is still assessed primarily through outcome-level metrics such as final-answer accuracy or formal proof verifiability, obscuring whether success reflects the discovery of the mathematical structure needed for a solution or merely the successful execution of a derivation.

Understanding mathematical reasoning therefore requires more than verifying that a derivation is correct; it requires exposing the ideas that organize the argument—why it works, which components are essential, and what can be generalized or reused. This need becomes increasingly salient as LLM-generated proofs grow in scale and complexity: lengthy chains of valid steps may establish correctness while leaving the underlying mathematical structure difficult to identify. Recent work has begun to emphasize precisely these aspects of mathematical reasoning, calling for representations that reveal the motivation, structure, and key ideas underlying an argument rather than only its final outcome~\citep{frieder2024data}. In his ICM 2026 lecture, Terence Tao similarly highlighted the importance of making machine-generated mathematics conceptually digestible~\citep{tao2026mathematics}, warning that optimizing primarily for formal correctness risks producing an increasing volume of valid but opaque proofs without a corresponding increase in mathematical understanding.
For LLMs, this perspective shifts the focus from merely producing valid proofs to determining whether the structural knowledge organizing those proofs is actually accessible to the model.

\begin{figure}[t]
    \centering
    \vspace{-0.15in}
    \includegraphics[width=\linewidth]{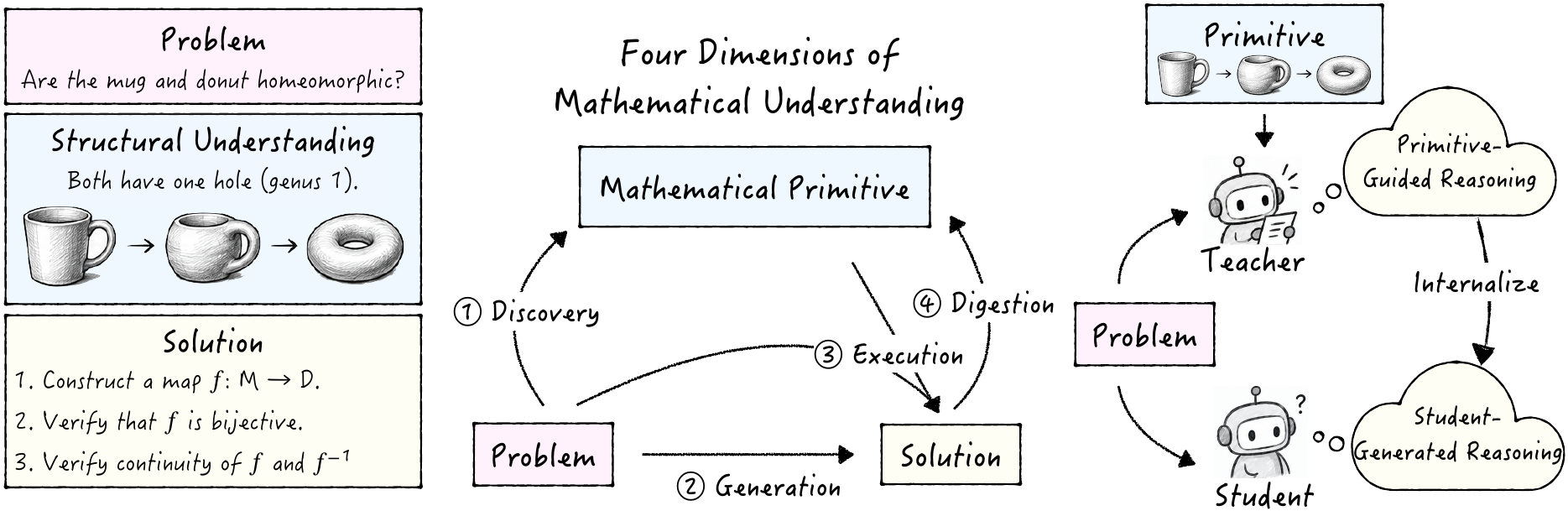}
    \vspace{-0.2in}
    \caption{\textbf{Diagnosing and internalizing mathematical primitives.} We introduce Mathematical Primitives to probe structural mathematical understanding in LLMs across four dimensions: Discovery, Generation, Digestion, and Execution. Our diagnosis further motivates \abs{}, which selectively transfers primitive-guided reasoning into the student without requiring primitives at inference time.}
    \vspace{-0.2in}
    \label{fig:teaser}
\end{figure}

In this paper, we systematically study structural mathematical understanding in LLMs, as illustrated in Figure~\ref{fig:teaser}: the ability to identify the compact, load-bearing mathematical structure that explains \textit{why a solution works}, distinct from the procedural ability to generate the derivation itself. We operationalize this notion through the \textbf{Mathematical Primitive}, a concise, non-procedural representation of the foundational idea organizing a solution, and introduce \hlei{}, a benchmark spanning four complementary dimensions: Discovery, Generation, Digestion, and Execution. Our diagnosis reveals that \ding{182} \textbf{similar solving accuracy can mask markedly different capability profiles}, \ding{183} \textbf{correct primitives unlock substantial latent execution capacity}, and, most importantly, \ding{184} \textbf{independent Discovery is the dominant bottleneck} to mathematical reasoning. We further find that discovery-limited failures are substantially more amenable to post-training, but existing methods can introduce regressions on problems the model already solves. Motivated by these findings, we introduce \abs{}, a post-training paradigm that uses Mathematical Primitives as privileged information and a bounded override mechanism to selectively transfer their guidance into the student's reasoning process, enabling primitive-guided reasoning to be internalized without access to primitives at inference time. Our main contributions are as follows:
\begin{itemize}[leftmargin=*,nosep]
    \item \textbf{Benchmarking structural mathematical understanding.} We introduce the notion of Mathematical Primitive and propose \hlei{}, a benchmark that evaluates mathematical reasoning across four complementary dimensions: Discovery, Generation, Digestion, and Execution.

    \item \textbf{Systematic Diagnosis.} Our analysis reveals distinct capability profiles hidden by answer accuracy, substantial latent execution capacity unlocked by correct primitives, and independent Discovery as the dominant bottleneck in mathematical reasoning. We further show that discovery-limited failures are substantially more amenable to post-training.

    \item \textbf{Internalizing primitive-guided reasoning.} Building on these findings, we introduce \abs{}, a primitive-privileged post-training paradigm that selectively transfers primitive-guided reasoning into the student model. Across multiple model scales and challenging mathematical reasoning benchmarks, \abs{} consistently improves average performance by \num{2.42} -- \num{4.48} points.
\end{itemize}

\section{Mathematical Primitives as Structural Understanding}\label{sec:primitive}

Rather than viewing mathematical problem solving as a sequence of steps, we separate \textbf{structural mathematical understanding} from \textbf{procedural execution}. We formalize the distinction through the \textbf{Mathematical Primitive} (\S\ref{subsec:primitive-def}) and introduce \hlei{}, a benchmark that evaluates LLM mathematical reasoning across four dimensions: Generation, Discovery, Digestion, and Execution (\S\ref{subsec:primitive-bench}).

\subsection{Mathematical Primitive}
\label{subsec:primitive-def}
A mathematical solution often begins before the first line of a proof, when an opaque problem is recognized through the structure that makes it solvable. We term this load-bearing structural understanding the \textbf{Mathematical Primitive} (Definition~\ref{def:primitive}). Unlike generic advice such as use ``induction'' or ``apply a theorem'', a primitive identifies the structural property that makes such a strategy applicable. It therefore serves as an intermediate reasoning object between problem understanding and procedural solution construction.
\begin{definition}[Mathematical Primitive]
\label{def:primitive}
For a mathematical problem, a \textbf{\emph{primitive}} is the essential conceptual observation that reveals \textbf{\emph{why}} the problem can be solved. It captures the latent mathematical structure shared by the problem and its solution, such as an invariant, a theorem condition, a representation, a reduction, or a reformulation. A primitive should be \textbf{\emph{concise}}, \textbf{\emph{problem-specific}}, and \textbf{\emph{explanatory}}: it must make clear what property is being exploited and how that property unlocks the solution. It is \textbf{\emph{not}} a routine calculation, a piece of generic advice, or a full proof.
\end{definition}


\subsection{Four Dimensions of Mathematical Understanding}
\label{subsec:four-dimension}


Taking the mathematical primitive as the fundamental concept of structural understanding, we systematically decompose mathematical reasoning into four complementary dimensions. Let $x$ denote the input problem, $p$ the corresponding mathematical primitive, $y$ a complete proof or solution, and $\pi$ an LLM solver. The four dimensions are defined as follows:
\begin{itemize}[leftmargin=*,nosep]

    \item \textbf{Discovery ($\pi(x) \rightarrow \hat{p}$)} measures whether the model can independently identify the load-bearing mathematical primitive from the input problem alone. 
    
    \item \textbf{Generation ($\pi(x) \rightarrow \hat{y}$)} evaluates whether the model can solve the problem directly under zero-shot CoT prompting, with performance measured by exact match on the final answer.

    \item \textbf{Digestion ($\pi(x,y) \rightarrow \hat{p}$)} instead provides the model with a complete correct solution and asks it to extract the mathematical primitive underlying the derivation, measuring whether the model can identify the underlying structure once a valid reasoning trajectory is exposed.

    \item \textbf{Execution ($\pi(x,p) \rightarrow \hat{y}$)} conditions the solver on the mathematical primitive and evaluates whether it can successfully solve the problem once the relevant structure is explicitly provided. 
\end{itemize}

Both Discovery and Digestion are evaluated using the scoring protocol introduced in \S~\ref{subsec:primitive-bench}. 

\subsection{The \hlei{} Benchmark}
\label{subsec:primitive-bench}

\textbf{Data Curation.}
We build on Humanity's Last Exam (HLE)~\citep{phan2025lastexam}, focusing on its mathematics, text-only, free-form subset with \texttt{exactMatch} evaluation. To reduce annotation noise in the original benchmark, we draw from the \textsc{HLE-Verified} release~\citep{zhai2026hle-verified}, retaining its \emph{Gold} and \emph{Revision} subsets, from which we randomly sample $200$ problems. For each problem, GPT-5.4-High generates an initial primitive annotation conditioned on the problem, reference answer, and gold rationale, following Definition~\ref{def:primitive}. Three human experts with graduate-level training in mathematics independently review the primitive and verify the validity of the problem, answer, and rationale. This process excludes $18$ unsuitable problems, yielding a final evaluation set of $182$. Further curation and annotation details are provided in Appendix~\ref{app:data-flilter}.


\textbf{Primitive Scoring.}
We evaluate each predicted primitive against the human-annotated gold primitive, using the reference solution only to recognize mathematically equivalent formulations. To account for different forms of mathematical structure, each problem is assigned to one of three primitive families, including \textsc{Recast}, \textsc{Witness}, and \textsc{Argument}, with full definitions provided in Appendix~\ref{app:primitive-family}.


Our scoring rubric evaluates three aspects of a predicted primitive: $V$ determines whether the response constitutes a valid primitive, $\sigma_{\text{gate}}$ measures whether it identifies the essential mathematical structure, and $\sigma_{\text{mech}}$ measures whether it explains how that structure enables the solution. We define
\begin{align*}
\mathrm{Score} = V \cdot \sigma_{\text{gate}}
\cdot
\big(0.6 + 0.4 \cdot \sigma_{\text{mech}}\big),
\quad
\mathrm{PrimitiveAcc} =  \mathbf{1}\left[\mathrm{Score}\ge\tau\right],
\end{align*}
with $\tau=0.8$. This formulation treats validity and structural identification as necessary conditions, while using mechanism quality to refine the score. Further details are provided in Appendix~\ref{app:annnotate}.


\section{Diagnosing Mathematical Reasoning}
\label{sec:dignos}
In this section, we conduct an empirical study of a diverse set of LLMs, including both open- and closed-source models, on \hlei{}. Moving beyond standard answer correctness, our evaluation systematically disentangles procedural execution from structural understanding. We report model performance across four dimensions: \textbf{Generation}, \textbf{Discovery}, \textbf{Digestion}, and \textbf{Execution}, following the evaluation protocol described in \S~\ref{subsec:primitive-bench}. The main results are summarized in Table~\ref{tab:digest-eval}.

\subsection{Evaluation Setup} 
\begin{wraptable}{r}{0.65\textwidth}
\vspace{-0.8\baselineskip}
\centering
\small
\setlength{\tabcolsep}{3pt}
\renewcommand{\arraystretch}{1.12}

\begin{tabular}{@{}lcccc@{}}
\toprule
\textbf{Model}
& \textbf{Discovery}
& \textbf{Generation}
& \makecell{\textbf{Digestion}\\[-2pt]
    {\scriptsize $\mathrm{Dig.}-\mathrm{Disc.}$}}
& \makecell{\textbf{Execution}\\[-2pt]
    {\scriptsize $\mathrm{Exec.}-\mathrm{Gen.}$}} \\
\midrule

\rowcolor{gray!8}[0pt][0pt]
\multicolumn{5}{@{}l}{\textit{OpenAI}} \\

gpt-5.4
& \num{82.42}
& \num{67.03}
& \num{100.00}\chg{+17.58}
& \num{86.81}\chg{+19.78} \\

gpt-5.4-mini
& \num{59.34}
& \num{50.55}
& \num{97.80}\chg{+38.46}
& \num{71.98}\chg{+21.43} \\

gpt-5.4-nano
& \num{41.76}
& \num{44.51}
& \num{97.25}\chg{+55.49}
& \num{68.68}\chg{+24.17} \\

gpt-oss-20B
& \num{34.62}
& \num{43.41}
& \num{91.21}\chg{+56.59}
& \num{60.99}\chg{+17.58} \\

\addlinespace[2pt]

\rowcolor{gray!8}[0pt][0pt]
\multicolumn{5}{@{}l}{\textit{Qwen}} \\

Qwen3.6-27B
& \num{24.73}
& \num{52.75}
& \num{92.31}\chg{+67.58}
& \num{78.57}\chg{+25.82} \\

Qwen3.5-27B
& \num{28.57}
& \num{47.80}
& \num{93.41}\chg{+64.84}
& \num{71.98}\chg{+24.18} \\

Qwen3.5-9B
& \num{13.19}
& \num{38.46}
& \num{79.67}\chg{+66.48}
& \num{61.54}\chg{+23.08} \\

Qwen3.5-4B
& \num{6.59}
& \num{26.37}
& \num{70.88}\chg{+64.29}
& \num{56.04}\chg{+29.67} \\

\addlinespace[2pt]

\rowcolor{gray!8}[0pt][0pt]
\multicolumn{5}{@{}l}{\textit{DeepSeek-R1 Distill}} \\

R1-0528-8B
& \num{6.59}
& \num{17.03}
& \num{68.68}\chg{+62.09}
& \num{39.56}\chg{+22.53} \\

R1-Distill-32B
& \num{6.04}
& \num{18.13}
& \num{65.38}\chg{+59.34}
& \num{43.41}\chg{+25.28} \\

R1-Distill-14B
& \num{4.40}
& \num{15.93}
& \num{67.03}\chg{+62.63}
& \num{39.01}\chg{+23.08} \\

R1-Distill-7B
& \num{7.14}
& \num{13.19}
& \num{48.90}\chg{+41.76}
& \num{32.42}\chg{+19.23} \\

\bottomrule
\end{tabular}

\caption{
Overall performance on the four dimensions of \hlei{}. Detailed results are presented in Table~\ref{tab:gap-family} in Appendix~\ref{app:add-diagnosis-exp}.
}
\label{tab:digest-eval}
\vspace{-0.8\baselineskip}
\end{wraptable}
We evaluate 12 models spanning three lineages and a broad capability range. The OpenAI lineage includes the closed-source GPT-5.4~\citep{gpt5.4}, GPT-5.4-Mini~\citep{gpt5.4-mini}, and GPT-5.4-Nano~\citep{gpt5.4-mini}, alongside the open-weight GPT-OSS-20B~\citep{openai2025gptoss120bgptoss20bmodel}. The Qwen lineage comprises Qwen-3.6-27B~\citep{qwen36_27b} and Qwen-3.5-\{27B, 9B, 4B\}~\citep{qwen35blog}. The DeepSeek lineage features DeepSeek-R1-0528-Qwen3-8B~\citep{deepseekai2025deepseekr1incentivizingreasoningcapability} and DeepSeek-R1-Distill-Qwen-\{32B, 14B, 7B\}~\citep{deepseekai2025deepseekr1incentivizingreasoningcapability}. We use high reasoning effort when available and a maximum generation budget of 120k tokens per problem.

\subsection{Results Analysis}\label{subsec:eval-res}
Table~\ref{tab:digest-eval} presents the comprehensive evaluation results of all models on \hlei{}. Performance in the Discovery and Digestion dimensions is assessed by the model's capacity to isolate the underlying mathematical primitive, utilizing the scoring protocol detailed in \S~\ref{subsec:primitive-bench}. Meanwhile, Generation and Execution are evaluated based on standard final-answer accuracy, following the HLE benchmark~\citep{phan2025lastexam}. Further details on the diagnostic evaluation and additional analyses are provided in Appendix~\ref{app:diagno-detail}.



\begin{figure}[ht]
    \centering
    \vspace{-0.1in}
    \includegraphics[width=\linewidth]{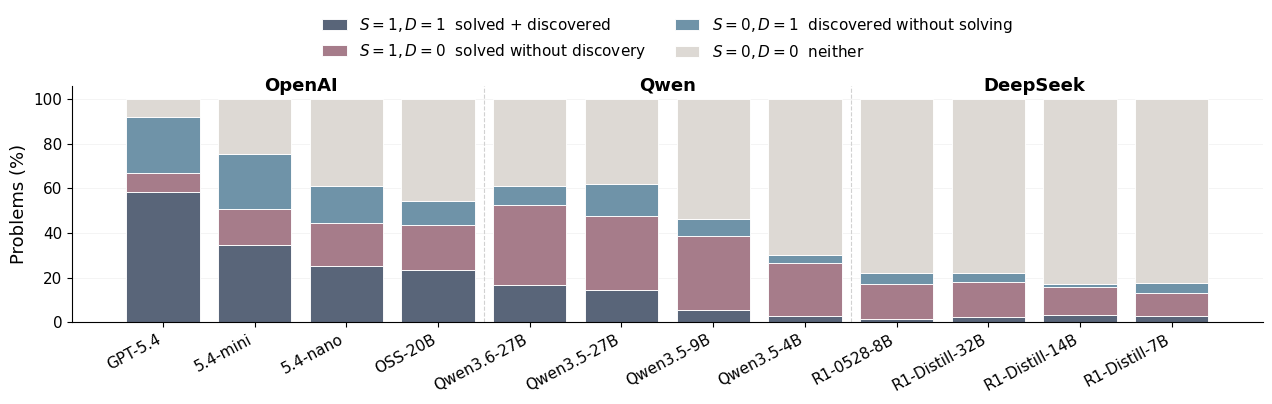}
    \vspace{-0.25in}
    \caption{Joint composition of Generation and Discovery outcomes across models.}
    \label{fig:s-d-composition}
    \vspace{-0.15in}
\end{figure}

\textbf{Finding 1: Answer accuracy masks distinct capability profiles.} 
Final-answer accuracy conflates structural discovery with downstream execution. As shown in Figure~\ref{fig:s-d-composition}, Discovery and Generation are positively associated overall, yet models with similar solving accuracy can exhibit sharply different structural capabilities. For example, Qwen3.6-27B and gpt-5.4-mini achieve comparable Generation performance, while differing by over $30$ percentage points in Discovery. Conversely, GPT-5.4 achieves substantially higher Discovery than Generation. The problem-level decomposition in Figure~\ref{fig:s-d-composition} reveals similarly distinct profiles across model families. The OpenAI lineage is more \textbf{primitive-forward}, with successful Generation frequently accompanied by correct Discovery and a substantial fraction of cases with correct Discovery but failed Generation ($D^+G^-$). In contrast, Qwen exhibits more cases with failed Discovery but successful Generation ($D^-G^+$), despite competitive Generation performance. The DeepSeek-R1 distilled models are more heavily concentrated in joint Discovery--Generation failures. Thus, similar top-line accuracy can arise from markedly different underlying capability profiles.


\begin{wrapfigure}{r}{0.4\textwidth}
    \centering
    \vspace{-0.1in}
    \includegraphics[width=\linewidth]{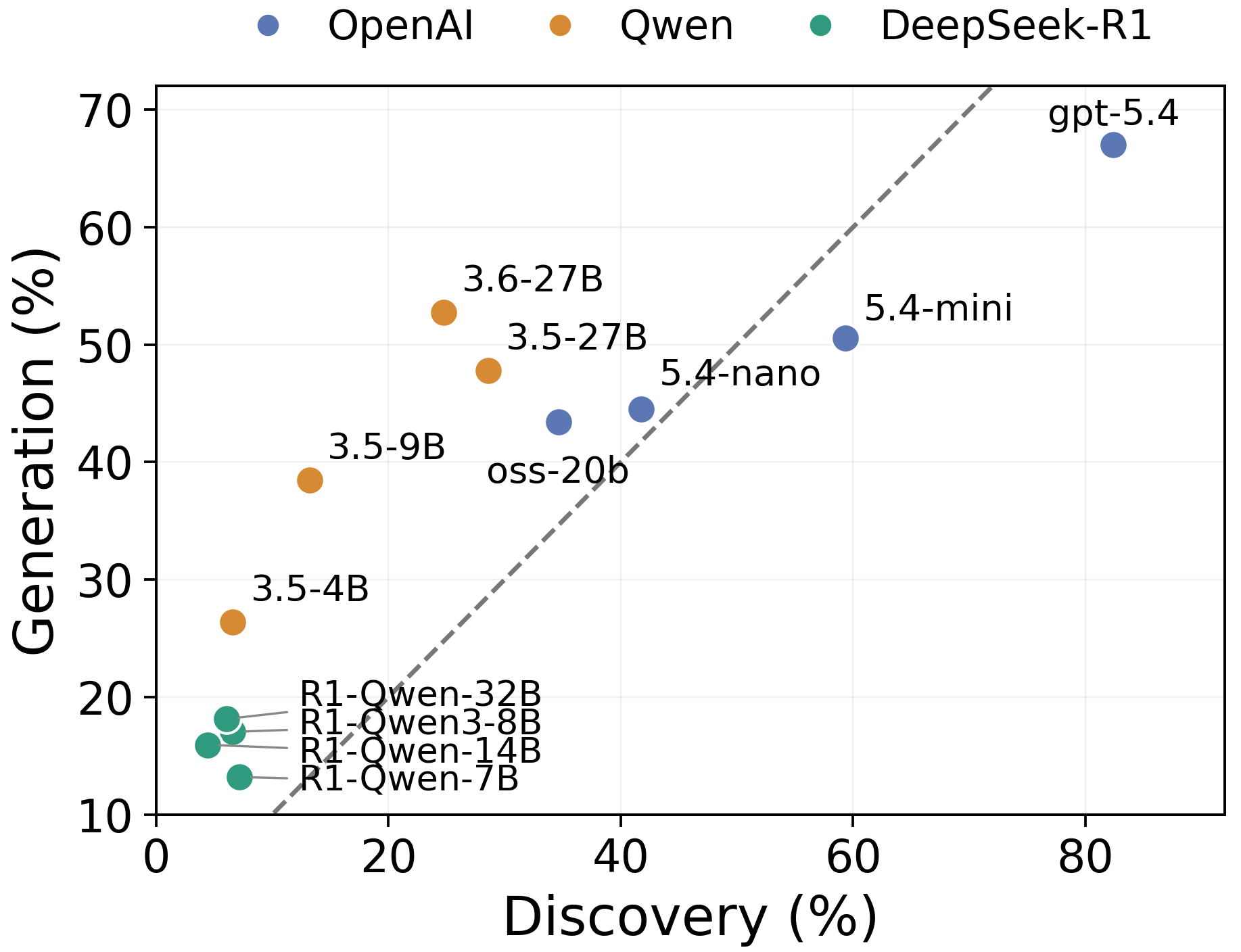}
    \vspace{-0.2in}
    \caption{
    Discovery versus Generation across all evaluated models on \hlei{}.
    }
    \label{fig:discovery-generation}
    \vspace{-0.12in}
\end{wrapfigure}

Qualitative analysis further clarifies these asymmetries. For gpt-5.4, $D^+G^-$ cases primarily reflect two post-primitive failure modes: \ding{182} \textbf{technical knowledge gaps}, where the correct structure is identified but a required lemma, theorem condition, or domain-specific fact is missing; and \ding{183} \textbf{procedural execution failures}, such as algebraic errors, missed edge cases, or unresolved logical gaps. 
For Qwen3.6-27B, among $D^-G^+$ cases, the decisive structural idea often emerges after an initial attempt without using it, typically around one quarter into the reasoning trace; in most remaining cases it appears at the outset, and only rarely is it absent entirely. The corresponding cold Discovery output often captures a partial form of the correct idea but misses its decisive component. Together with the longer reasoning traces in Figure~\ref{fig:token-distributions}, this pattern is consistent with a more \textbf{grind-first} profile, where the relevant structure tends to \textbf{emerge during extended procedural exploration} rather than being isolated before execution.

\textbf{Finding 2: Correct primitives unlock latent execution capacity.}
To examine whether models possess downstream capabilities that are not realized during direct Generation, we evaluate Execution when the correct Mathematical Primitive is explicitly provided. As shown in Table~\ref{tab:digest-eval}, primitive guidance improves performance by $17.58$ to $29.67$ percentage points across all 12 models. Notably, under the same primitive-guided setting, the Qwen models achieve Execution performance competitive with the closed-source gpt-5.4 variants despite substantially weaker Discovery. This gap shows that difficulty independently identifying the relevant mathematical structure can coexist with strong downstream problem-solving capability once that structure is available. More broadly, these gains indicate that many Generation failures mask downstream solving capacity that models can successfully exercise when the relevant mathematical structure is supplied. 


\sisetup{
    detect-family = true,
    detect-weight = true
}
\vspace{-0.1in}
\begin{table}[htbp]
\centering
\footnotesize
\setlength{\tabcolsep}{3pt}
\renewcommand{\arraystretch}{1.15}

\begin{tabular}{@{}lccccc@{}}
\toprule
\multirow{2}{*}{\textbf{Model}}
& \multirow{2}{*}{\textbf{Generation}}
& \multicolumn{4}{c}{\textbf{Execution}} \\
\cmidrule(lr){3-6}

&
& {\textbf{Self-Generated Primitive}}
& {\textbf{Teacher Plan}}
& {\textbf{Teacher Primitive}}
& {\textbf{Gold Primitive}} \\
\midrule

gpt-5.4-mini
& \num{50.55}
& \num{48.90}\chg{-1.65}
& \num{60.44}\chg{+9.89}
& \num{63.74}\chg{+13.19}
& \num{71.98}\chg{+21.43} \\

gpt-5.4-nano
& \num{44.51}
& \num{46.15}\chg{+1.64}
& \num{54.40}\chg{+9.89}
& \num{60.99}\chg{+16.48}
& \num{68.68}\chg{+24.17} \\

Qwen3.6-27B
& \num{52.75}
& \num{41.21}\chg{-11.54}
& \num{56.04}\chg{+3.29}
& \num{64.84}\chg{+12.09}
& \num{78.57}\chg{+25.82} \\

\bottomrule
\end{tabular}
\vspace{-0.1in}
\caption{
Generation and Execution performance on \hlei{} under different forms of
reasoning support. Detailed results are presented in Table \ref{tab:execution-primitive-app} in Appendix \ref{app:add-diagnosis-exp}.
}
\label{tab:execution-primitive}
\vspace{-0.1in}
\end{table}

We further investigate whether this rescue can be attributed simply to task decomposition or generic procedural scaffolding. Table~\ref{tab:execution-primitive} compares four intervention settings. Conditioning the model on its own generated primitive provides little benefit: it yields only a marginal gain for gpt-5.4-nano and substantially degrades Qwen3.6-27B, indicating that a forced discover-then-execute decomposition alone is insufficient. We then compare two forms of external guidance generated by the same teacher model (gpt-5.4) from the problem alone: a step-by-step \textbf{Teacher Plan}, which serves as a procedural hint by specifying the major solution steps, and a \textbf{Teacher Primitive}, which is generated according to the definition of primitive. Although the Teacher Plan improves over direct Generation, the Teacher Primitive yields substantially larger gains across the evaluated models. The Gold Primitive further maximizes this effect, pushing accuracy to $71.98\%$, $68.68\%$, and $78.57\%$ for the representative models. This contrast provides crucial empirical validation for our premise in \S\ref{subsec:primitive-def}: a mathematical primitive is fundamentally distinct from, and vastly more effective than, a generic plan or hint. The rescue effect therefore cannot be attributed to task decomposition or the procedural scaffold evaluated here alone; exposing the load-bearing mathematical structure provides a substantially stronger signal for downstream reasoning.

\textbf{Finding 3: Independent structural discovery is the dominant bottleneck.}
As shown in Table~\ref{tab:digest-eval}, models perform substantially better on \textbf{Digestion}, recovering the Mathematical Primitive from a correct solution, than on \textbf{Discovery}, where the same primitive must be identified from the problem alone. 
This asymmetry is particularly pronounced outside the strongest OpenAI models. Even models with weak Discovery can often recover the organizing primitive once the relevant reasoning is exposed through a completed solution; for instance, Qwen3.6-27B rises from 24.73\% Discovery to 92.31\% Digestion and R1-Distill-32B from 6.04\% to 65.38\%. 
\begin{wrapfigure}{r}{0.25\textwidth}
    \centering
    \vspace{-0.8\baselineskip}
    \includegraphics[width=\linewidth]{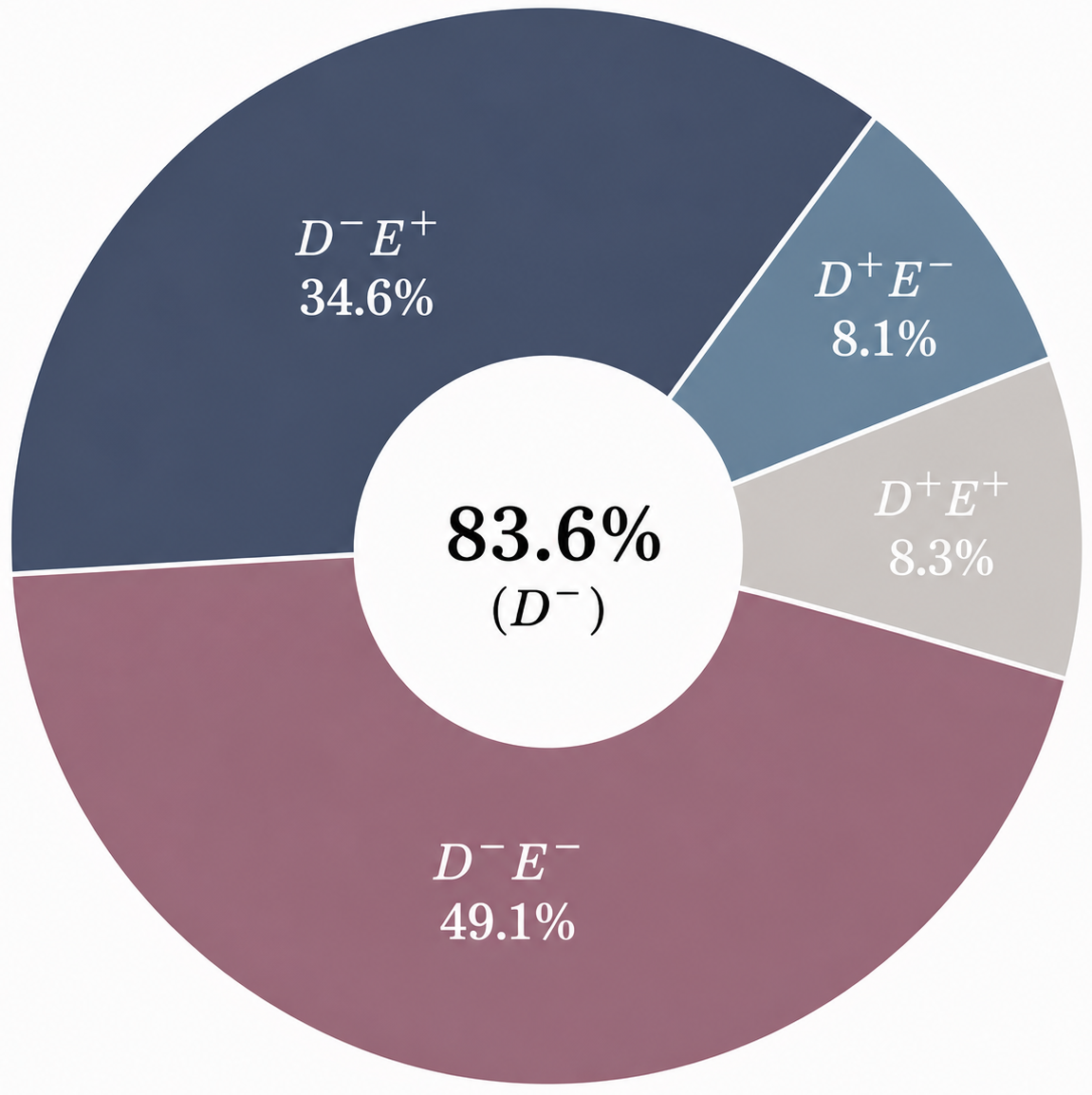}
    \vspace{-0.2in}
    \caption{Decomposition of Generation failures by Discovery ($D$) and Execution ($E$). Detailed results can be found in Table~\ref{tab:failure-composition} in Appendix~\ref{app:add-diagnosis-exp}.}
    \label{fig:failure-decomposition}
    \vspace{-4em}
\end{wrapfigure}
These results reveal a pronounced gap between retrospective recognition and prospective discovery of mathematical structure: models can often recognize the key idea once it is instantiated in a valid solution, yet struggle to identify it independently before the derivation is known.

To further localize this bottleneck, we decompose Generation failures according to the joint outcomes of Discovery ($D$) and Execution ($E$), as shown in Figure~\ref{fig:failure-decomposition}. A striking $83.6\%$ of all failures fall into the $D^{-}$ regime, indicating that failed primitive discovery dominates the failure landscape. Importantly, a substantial fraction of these cases are \textbf{discovery-limited} failures ($D^{-}E^{+}$): the model fails to identify the correct primitive independently, yet succeeds once it is supplied. This reveals that, for many failures, latent execution capacity is already present but remains inaccessible because of the upstream bottleneck in primitive discovery.

\subsection{Further Analysis: Which Failures Are Repairable?}
\label{subsec:diagnosis-repair}
While our preceding diagnosis reveals the failure modes of LLMs during inference, a natural follow-up question arises: \textbf{how does post-training impact these mathematical reasoning failures?} Of particular interest are \textbf{discovery-limited} failures, which exhibit substantial room for improvement (\S\ref{subsec:eval-res}). We hypothesize that these failures lie just at the boundary of the models' reasoning capabilities, making them highly amenable to repair through post-training. Therefore, we utilize Supervised Fine-Tuning (SFT) and On-Policy Self-Distillation (OPSD)~\citep{zhao2026opsd} to investigate how different failure modes uniquely respond to these post-training paradigms.

\textbf{Setup.} 
Our training corpus contains 709 Mathematics Ph.D.\ qualifying-examination problems from 1991--2026, each paired with a human-authored proof (median length: 118 words); additional details are provided in Appendix~\ref{app:dataset-details}. We use qualifying-examination problems because their open-ended, proof-oriented nature closely matches the difficulty and structural reasoning demands of \hlei{}. Since such problems generally lack algorithmically verifiable short answers, we focus on text-supervised post-training rather than outcome-verified methods such as RLVR. We evaluate SFT and OPSD on Qwen-3.5-{27B, 9B, 4B}~\citep{qwen35blog}, using the default OPSD configuration, bfloat16 precision, a fixed seed of $42$, and NVIDIA RTX 6000 Ada GPUs.

\textbf{On-Policy Self-Distillation.}
Given training examples $\{x,y\}$, where $x$ denotes the input problem and $y$ its reference solution, both the teacher and student policies are initialized from the same base model $\pi_0$. At each decoding step $t$, the student autoregressively generates an on-policy response, inducing the token distribution $P_t := \pi_S(\cdot \mid \hat{y}_{<t}, x)$, while the teacher distribution $Q_t := \pi_T(\cdot \mid \hat{y}_{<t}, x, y)$ additionally conditions on the reference solution. Notably, both policies are conditioned on the same student prefix $\hat{y}_{<t}$, allowing the teacher to provide token-level supervision along the student's own reasoning trajectory. OPSD minimizes the divergence between these distributions:
\begin{align*}
\mathcal{L}_{\mathrm{OPSD}} = \frac{1}{T}\sum_{t=1}^{T}
D(P_t, Q_t) = \frac{1}{T} \sum_{t=1}^{T} D\left(\pi_S(\cdot \mid \hat{y}_{<t}, x), \pi_T(\cdot \mid \hat{y}_{<t}, x,y) \right),
\end{align*}
where $D$ is instantiated as Kullback--Leibler (KL), reverse KL (rKL), or Jensen--Shannon (JSD). Specifically, KL uses $D_{\mathrm{KL}}(Q_t\parallel P_t)$, rKL uses $D_{\mathrm{KL}}(P_t\parallel Q_t)$, and JSD is symmetric in $P_t$ and $Q_t$.

\textbf{Results.} 
Across the 27B, 9B, and 4B models, the base models leave $341$ failure instances in \hlei{}. We focus on the $294$ cases with failed Discovery: $147$ $D^-E^+$ and $147$ $D^-E^-$ cases. We define repair as an incorrect baseline Generation response becoming correct after post-training. As shown in Table~\ref{tab:repair-quadrants}, SFT and OPSD repair $20.4\%$ and $21.1\%$ of $D^-E^+$ cases, roughly three 
\begin{wraptable}{r}{0.33\textwidth}
\centering
\small
\vspace{-0.15in}
\setlength{\tabcolsep}{6pt}
\begin{tabular}{lrr}
\toprule
\textbf{Failure type} & \textbf{SFT} & \textbf{OPSD} \\
\midrule
$D^-E^+$
& 20.4\%
& 21.1\% \\
$D^-E^-$
& 6.8\%
& 8.2\% \\
\bottomrule
\end{tabular}
\vspace{-0.1in}
\caption{Post-training repair rates of SFT and OPSD across initial \hlei{} failure quadrants.}
\label{tab:repair-quadrants}
\vspace{-0.8\baselineskip}
\end{wraptable}
times their respective rates of $6.8\%$ and $8.2\%$ for $D^-E^-$ cases. 
This ordering holds at each model scale, identifying discovery-limited failures as a promising target for post-training. However, net gains in Generation remain modest (as shown in Table~\ref{tab:ipsd-results}) because some initially correct cases become incorrect, offsetting part of these repairs. These regressions are consistent with post-training drift and suggest that supervision from reference proofs should be transferred more selectively.

\section{Internalizing Primitive-Guided Reasoning}
\label{sec:internalize}
In this section, we introduce \abs{}, a novel variant of on-policy self-distillation that improves the mathematical problem-solving capability of LLMs by internalizing primitive-guided reasoning, motivated by the observations in \S\ref{sec:dignos}. Extensive experiments with Qwen models as backbones show that \abs{} consistently outperforms strong baselines on a diverse set of mathematical reasoning benchmarks (\S\ref{subsec:ipsd-exp}). Further implementation details and ablation results are provided in Appendix~\ref{app:ablations}.


\subsection{Primitives as Privileged Information} 

\begin{wrapfigure}{l}{0.7\textwidth}
    \centering
    \vspace{-0.1in}
    \includegraphics[width=\linewidth]{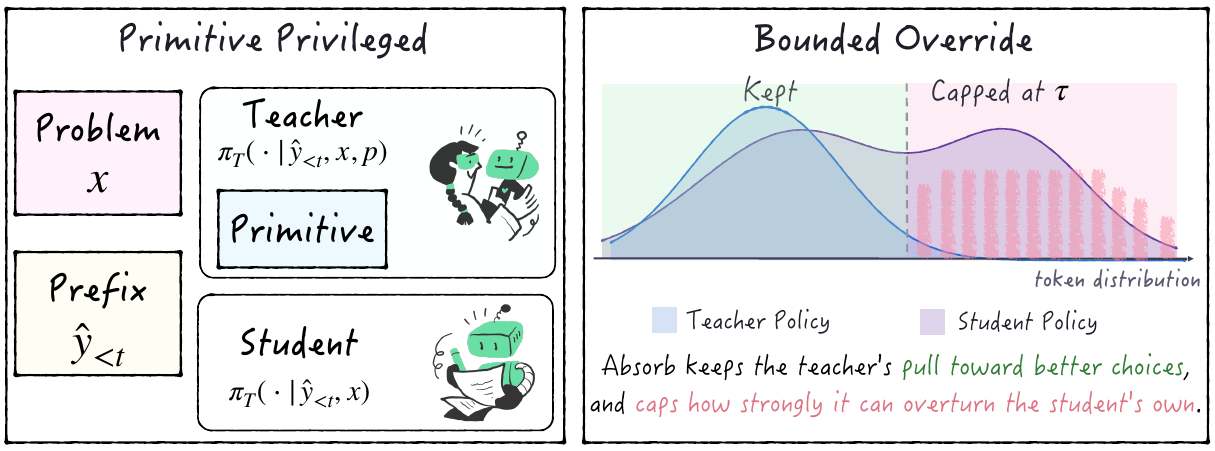}
    \vspace{-0.2in}
    \caption{Illustration of \abs{} paradigm.}
    \label{fig:ipsd-illustration}
    \vspace{-0.15in}
\end{wrapfigure}
Our diagnosis in \S\ref{sec:dignos} identifies missing mathematical primitives as the primary bottleneck in discovery-limited problems. A natural remedy is to directly supervise primitive generation. In practice, however, this objective transfers poorly to solution reasoning, likely reflecting a mismatch between structural understanding and step-by-step generation, as further supported by the ablations in Appendix~\ref{app:ablations}. More importantly, \S\ref{subsec:diagnosis-repair} shows that post-training can repair discovery-limited failures while leaving Discovery largely unchanged. This dissociation suggests that effective reasoning does not require primitives to be explicitly generated; instead, they can guide the reasoning trajectory implicitly. 


Motivated by this observation, we provide \textbf{mathematical primitives as privileged information} to the teacher rather than as prediction targets for the student. This mirrors our \textbf{Execution} evaluation, where providing the primitive unlocks latent execution capability, and uses the same primitive-conditioned reasoning to guide post-training. Compared with full reference proofs, primitives provide more targeted supervision by supplying the missing structural information without prescribing execution steps the student can already perform. This design transfers \textbf{primitive-guided reasoning} into the student's own generation process without requiring primitives at inference time.

\subsection{Bounded Override}
Given the reasoning drift observed with existing post-training methods, indiscriminate teacher intervention should be avoided. Moreover, introducing mathematical primitives as privileged information can cause substantial privilege leakage under naive OPSD on complex reasoning problems, as shown in Appendix~\ref{app:ablations}. This motivates a selective transfer mechanism that preserves useful privileged guidance while limiting how strongly it can override the student's own reasoning. Intuitively, the teacher should \textbf{pull the student toward better choices without abruptly displacing preferences} that the student has already formed. The primitive should refine the student's existing reasoning space rather than force it to imitate decisions that are only accessible under privileged context. 


This leads to two design principles: \ding{182} when the privileged teacher assigns higher preference to a choice than the student, this positive guidance should be transferred in full, as it directly reflects information contributed by the primitive; and \ding{183} when the teacher strongly suppresses a choice favored by the student, the resulting pressure should be bounded, since large disagreement may arise from privileged information unavailable to the student rather than from an error it can independently recognize. Importantly, bounding this negative pressure does not freeze incorrect student preferences: positive guidance toward better alternatives remains fully active, progressively shifting probability mass away from suboptimal choices over training. These desiderata admit a simple implementation through reverse KL with one-sided clamping, which preserves positive teacher guidance while bounding excessive suppression of student-preferred choices.

\begin{table}[t]
\centering
\vspace{-0.1in}
\small
\setlength{\tabcolsep}{3pt}
\renewcommand{\arraystretch}{1.2}
\begin{tabular}{@{}lllllll@{}}
\xheavyseven
& \multicolumn{2}{c}{\textbf{\hlei{}}}
& \multirow{2}{*}{\textbf{HLE Math}}
& \multirow{2}{*}{\textbf{HMMT25}}
& \multirow{2}{*}{\textbf{Omni-MATH}}
& \multirow{2}{*}{\textbf{Avg.}} \\
\cmidrule(lr){2-3}

\textbf{Model}
& \textbf{Discovery}
& \textbf{Generation}
& & & & \\

\xlightseven

Qwen3.5-4B
& \num{6.59}
& \num{26.37}
& \num{26.70}
& \num{76.67}
& \num{78.00}
& \num{51.94} \\

\;+ SFT
& \num{5.49}\chg{-1.10}
& \num{25.27}\chg{-1.10}
& \num{27.36}\chg{+0.65}
& \num{83.33}\chg{+6.66}
& \num{79.33}\chg{+1.33}
& \num{53.82}\chg{+1.89} \\

\;+ OPSD
& \num{8.79}\chg{+2.20}
& \num{30.22}\chg{+3.85}
& \num{29.58}\chg{+2.88}
& \num{83.33}\chg{+6.66}
& \num{79.33}\chg{+1.33}
& \num{55.62}\chg{+3.68} \\

\rowcolor{gray!8}
\;+ \textbf{\abs{}}
& {\bfseries\num{9.89}}\chg{+3.30}
& {\bfseries\num{31.32}}\chg{+4.95}
& {\bfseries\num{31.02}}\chg{+4.32}
& \num{83.33}\chg{+6.66}
& {\bfseries\num{80.00}}\chg{+2.00}
& {\bfseries\num{56.42}}\chg{+4.48} \\

\xlightseven

Qwen3.5-9B
& \num{13.19}
& \num{38.46}
& \num{35.60}
& \num{90.00}
& \num{78.67}
& \num{60.68} \\

\;+ SFT
& \num{12.64}\chg{-0.55}
& \num{32.42}\chg{-6.04}
& \num{34.55}\chg{-1.05}
& \num{90.00}\chg{+0.00}
& \num{80.67}\chg{+2.00}
& \num{59.41}\chg{-1.27} \\

\;+ OPSD
& \num{10.99}\chg{-2.20}
& \num{33.52}\chg{-4.95}
& \num{35.08}\chg{-0.52}
& \num{90.00}\chg{+0.00}
& \num{81.33}\chg{+2.67}
& \num{59.98}\chg{-0.70} \\

\rowcolor{gray!8}
\;+ \textbf{\abs{}}
& {\bfseries\num{14.29}}\chg{+1.10}
& {\bfseries\num{43.96}}\chg{+5.49}
& {\bfseries\num{38.35}}\chg{+2.75}
& {\bfseries\num{93.33}}\chg{+3.33}
& {\bfseries\num{82.00}}\chg{+3.33}
& {\bfseries\num{64.41}}\chg{+3.73} \\

\xlightseven

Qwen3.5-27B
& {\bfseries\num{28.57}}
& \num{47.80}
& \num{42.67}
& \num{96.67}
& \num{87.33}
& \num{68.62} \\

\;+ SFT
& \num{28.02}\chg{-0.55}
& \num{48.35}\chg{+0.55}
& \num{43.59}\chg{+0.92}
& \num{96.67}\chg{+0.00}
& \num{86.67}\chg{-0.67}
& \num{68.82}\chg{+0.20} \\

\;+ OPSD
& {\bfseries\num{28.57}}\chg{+0.00}
& \num{44.51}\chg{-3.30}
& \num{42.80}\chg{+0.13}
& \num{96.67}\chg{+0.00}
& \num{88.00}\chg{+0.67}
& \num{67.99}\chg{-0.62} \\

\rowcolor{gray!8}
\;+ \textbf{\abs{}}
& \num{28.02}\chg{-0.55}
& {\bfseries\num{48.90}}\chg{+1.10}
& {\bfseries\num{46.60}}\chg{+3.93}
& {\bfseries\num{100.00}}\chg{+3.33}
& {\bfseries\num{88.67}}\chg{+1.33}
& {\bfseries\num{71.04}}\chg{+2.42} \\

\xheavyseven

\end{tabular}
\vspace{-0.1in}
\caption{
Comparison of SFT, OPSD, and \abs{} across three Qwen3.5 model scales. 
}
\label{tab:ipsd-results}
\vspace{-0.2in}
\end{table}

\subsection{Overall Objective}
Combining the two design principles above, we introduce \abs{}, illustrated in Figure~\ref{fig:ipsd-illustration}. \abs{} conditions the teacher on the mathematical primitive $p$ to provide targeted guidance for the missing discovery signal, while bounding how strongly this can override the student's own reasoning. Let $\mathcal{S}_t$ denote the teacher's top-$K$ token support at step $t$, and let $\bar{\pi}_S$ and $\bar{\pi}_T$ denote the student and privileged-teacher distributions renormalized over $\mathcal{S}_t$, respectively. The resulting objective is:
\begin{align*}
\mathcal{L}_{\abs{}} = \frac{1}{T}\sum_{t=1}^{T} \sum_{v\in\mathcal{S}_t} \min\left\{ \bar{\pi}_S(v \mid \hat{y}_{<t}, x) \log \frac{\bar{\pi}_S(v \mid \hat{y}_{<t}, x)}
{\bar{\pi}_T(v \mid \hat{y}_{<t}, x, p)}, \tau \right\}.
\label{eq:absorb-loss}
\end{align*}
Here, the privileged teacher injects primitive-guided information along the student's on-policy reasoning trajectory, while the one-sided clamp preserves its positive guidance toward better alternatives and bounds excessive suppression of choices favored by the student.



\subsection{Experimental Results}
\label{subsec:ipsd-exp}
\textbf{Setup.}
For implementing \abs{}, we utilize the identical base models and optimization configurations detailed in \S\ref{subsec:diagnosis-repair}, differing only in the substitution of the standard OPSD objective with our primitive-privileged divergence loss ($\mathcal{L}_{\text{ABSORB}}$). To comprehensively evaluate internalized reasoning capabilities and generalization, we assess performance across four challenging mathematical benchmarks: our curated \hlei{} (evaluating both Generation and Discovery), the full mathematics subset of HLE-Verified~\citep{zhai2026hle-verified}, HMMT25~\citep{dekoninck2026hmmt}, and Omni-MATH~\citep{gao2025omni}. To account for the extreme difficulty of the competition-level datasets, we report pass@4 accuracy for HMMT25 and Omni-MATH. The overall average (Avg.) is computed exclusively over the Generation scores across all four benchmarks, excluding Discovery.

\textbf{Results.}
As shown in Table~\ref{tab:ipsd-results}, \abs{} consistently outperforms SFT and OPSD across model scales and benchmarks. At the 9B scale, SFT and OPSD reduce Generation accuracy by $6.04$ and $4.95$ percentage points, respectively, while OPSD incurs a $3.30$-point drop at the 27B scale. These regressions explain the modest net gains discussed in \S\ref{subsec:diagnosis-repair}: repairing previously failed problems can be offset by newly introduced errors on problems the model already solves. In contrast, \abs{} improves Generation at all three model scales, including a $5.49$ point gain at 9B and a $3.73$ point improvement in average benchmark performance. This pattern is consistent with the role of bounded override: selective transfer retains useful primitive-guided information while limiting disruptive changes to the model's existing reasoning behavior. The results also support the intended internalization of primitive-guided reasoning. Improvements from \abs{} are concentrated in downstream reasoning rather than explicit Discovery. At the 9B scale, Generation improves by $5.49$ points while Discovery increases by only $1.10$; at 27B, Generation improves by $1.10$ points and HLE Math by $3.93$, despite a $0.55$-point decrease in Discovery. Thus, the benefit of the primitive need not manifest as better explicit primitive recovery; instead, its structural guidance is increasingly reflected in the model's unassisted reasoning process.

\section{Related Work}

\paragraph{Mathematical Reasoning in LLMs.}
Mathematical reasoning benchmarks have rapidly escalated in difficulty, progressing from early grade-school and competition datasets (e.g., MathQA~\citep{amini2019mathqa}, ASDiv~\citep{miao-etal-2020-diverse}, SVAMP~\citep{patel2021nlp}, GSM8K~\citep{cobbe2021training}, MATH~\citep{hendrycks2021measuring}) to advanced Olympiad- and expert-level frontiers (TheoremQA~\citep{chen2023theoremqa}, MathBench~\citep{liu2024mathbench}, AIME~\citep{dekoninck2026hmmt}, OlympiadBench~\citep{he2024olympiadbench}, Omni-MATH~\citep{gao2025omni}, HLE~\citep{phan2025lastexam}, FrontierMath~\citep{glazer2024frontiermath}). Despite this scaling in complexity, evaluation predominantly remains rigidly anchored to final-answer correctness. Recognizing the limitations of outcome-based metrics, recent literature explores process-oriented evaluation: SMART~\citep{hou-etal-2026-smart} decomposes problem-solving across cognitive dimensions, ReasoningMath-Plus~\citep{zheng2026unmasking} verifies minimal procedural skeletons, and DeepInsightTheorem~\citep{li2026learning} leverages proof sketches for intermediate supervision. Executable Functional Abstractions (EFAs)~\citep{khan2025executable} represent mathematical structure as executable generative programs, whereas our Mathematical Primitive captures the concise, non-procedural idea underlying a specific solution. Our approach instead isolates the load-bearing \emph{mathematical primitive}, decoupling structural understanding from procedural execution. We evaluate whether models can discover it from a problem, recover it from a solution, and execute the proof when it is provided.

\paragraph{On-Policy Self-Distillation.}
On-Policy Self-Distillation (OPSD) advances knowledge distillation for LLM reasoning by mitigating the distribution shift inherent in traditional off-policy methods~\citep{zhao2026opsd,he2026self,penaloza2026privileged}. Specifically, the student generates its own reasoning trajectories on-policy, while the teacher, conditioned on additional privileged information, provides dense supervision along these rollouts, with both sharing the same base model. However, recent studies show that naive OPSD can suffer from privilege leakage and suppress exploratory or self-corrective reasoning. To address this, \citet{shen2026purified} isolate inference-transferable supervision from reference-induced shortcuts, while \citet{kaur2026rethinking,peng2026diagnosing} focus on preserving high-entropy decision points and epistemic behaviors from aggressive privileged-teacher corrections. Other approaches modulate supervision based on teacher--student disagreement~\citep{li2026demopsd,yu2026mismatch,xie2026trust}, or utilize RL primarily for token-level credit assignment~\citep{yang2026self}. AVSD mitigates privilege leakage by separating shared and view-specific signals across multiple privileged views~\citep{nguyen2026avsd}. \abs{} differs primarily in the privileged signal itself. Rather than conditioning the teacher on a full solution, \abs{} uses the primitive as privileged information and selectively transfers its guidance through a bounded override mechanism.

\section{Conclusion}

In this paper, we introduced the notion of Mathematical Primitive and \hlei{}, a benchmark for systematically evaluating structural mathematical understanding in LLMs across Discovery, Generation, Digestion, and Execution. Our diagnosis reveals that similar solution accuracy can mask markedly different capability profiles, correct primitives can unlock substantial latent execution capacity, and independent Discovery constitutes a dominant bottleneck in mathematical reasoning. We further show that discovery-limited failures are substantially more amenable to post-training, while existing methods can introduce regressions on problems the model already solves. Building on these findings, we proposed \abs{}, a primitive-privileged post-training paradigm that selectively transfers primitive-guided reasoning through a bounded override mechanism. Experiments on challenging mathematical reasoning benchmarks show that \abs{} consistently improves over strong post-training baselines. Overall, our results highlight the value of explicitly diagnosing structural mathematical understanding and leveraging this structure to improve LLM reasoning.

By systematically investigating mathematical understanding in LLMs, we hope this work encourages a broader shift in AI for Science from optimizing benchmark performance alone toward identifying the capabilities required for scientific problem solving and discovery. Given the foundational role of mathematics across STEM disciplines, this structural perspective may extend beyond mathematics. We highlight three promising directions. First, our diagnostic framework could be generalized to other structurally rigorous domains, such as theoretical physics and computational chemistry. Second, the observed discovery-execution asymmetry suggests a potential design principle for mathematical agents: explicitly separating structural search from procedural execution may enable more effective allocation of inference-time computation. Finally, future work could explore integrating mathematical primitives with formal verification systems such as Lean, connecting informal structural reasoning with formal proof generation.

\bibliography{iclr2027_conference}

@article{phan2025lastexam,
      title = {A benchmark of expert-level academic questions to assess {AI} capabilities},
      author = {{Center for AI Safety} and {Scale AI} and {HLE Contributors Consortium}},
      journal = {Nature},
      volume = {649},
      pages = {1139--1146},
      year = {2026},
      doi = {10.1038/s41586-025-09962-4},
      eprint = {2501.14249},
      archivePrefix = {arXiv},
      primaryClass = {cs.LG},
      url = {https://arxiv.org/abs/2501.14249}
}

@article{zhai2026hle-verified,
  title={HLE-Verified: A Systematic Verification and Structured Revision of Humanity's Last Exam},
  author={Zhai, Weiqi and Wang, Zhihai and Wang, Jinghang and Yang, Boyu and Li, Xiaogang and Xu, Xander and Wang, Bohan and Wang, Peng and Wu, Xingzhe and Li, Anfeng and others},
  journal={arXiv preprint arXiv:2602.13964},
  year={2026}
}

@misc{gpt5.4,
  author       = {OpenAI},
  title        = {Introducing GPT‑5.4},
  year         = {2026},
  url          = {https://openai.com/index/introducing-gpt-5-4/},
  month        = {March},
}

@misc{gpt5.4-mini,
  author       = {OpenAI},
  title        = {Introducing GPT‑5.4 mini and nano},
  year         = {2026},
  url          = {https://openai.com/index/introducing-gpt-5-4-mini-and-nano/},
  month        = {March},
}

@misc{openai2025gptoss120bgptoss20bmodel,
      title={gpt-oss-120b \& gpt-oss-20b Model Card}, 
      author={OpenAI},
      year={2025},
      eprint={2508.10925},
      archivePrefix={arXiv},
      primaryClass={cs.CL},
      url={https://arxiv.org/abs/2508.10925}, 
}

@misc{qwen36_27b,
    title = {{Qwen3.6-27B}: Flagship-Level Coding in a 27B Dense Model},
    url = {https://qwen.ai/blog?id=qwen3.6-27b},
    author = {{Qwen Team}},
    month = {April},
    year = {2026}
}

@misc{qwen35blog,
    title = {Qwen3.5: Accelerating Productivity with Native Multimodal Agents},
    url = {https://qwen.ai/blog?id=qwen3.5},
    author = {{Qwen Team}},
    month = {February},
    year = {2026}
}

@misc{deepseekai2025deepseekr1incentivizingreasoningcapability,
      title={DeepSeek-R1: Incentivizing Reasoning Capability in LLMs via Reinforcement Learning}, 
      author={DeepSeek-AI},
      year={2025},
      eprint={2501.12948},
      archivePrefix={arXiv},
      primaryClass={cs.CL},
      url={https://arxiv.org/abs/2501.12948}, 
}

@misc{openai2026tenadvances,
  title={Ten Advances in Mathematics and Theoretical Computer Science},
  author={{OpenAI}},
  year={2026},
  howpublished={\url{https://openai.com/index/ten-advances-in-mathematics/}},
  note={Astra Model Research Release and Lean 4 Formalizations}
}

@article{boye2025large,
  title={Large language models and mathematical reasoning failures},
  author={Boye, Johan and Moell, Birger},
  journal={arXiv preprint arXiv:2502.11574},
  year={2025}
}

@inproceedings{lu2024proof,
  title={Proof automation with large language models},
  author={Lu, Minghai and Delaware, Benjamin and Zhang, Tianyi},
  booktitle={Proceedings of the 39th IEEE/ACM International Conference on Automated Software Engineering},
  pages={1509--1520},
  year={2024}
}

@article{qin2025decomposing,
  title={Decomposing Elements of Problem Solving: What" Math" Does RL Teach?},
  author={Qin, Tian and Park, Core Francisco and Kwun, Mujin and Walsman, Aaron and Malach, Eran and Anand, Nikhil and Tanaka, Hidenori and Alvarez-Melis, David},
  journal={arXiv preprint arXiv:2505.22756},
  year={2025}
}

@article{frieder2024data,
  title={Data for mathematical copilots: Better ways of presenting proofs for machine learning},
  author={Frieder, Simon and Bayer, Jonas and Looi, Sam and Loader, Jacob and Berner, Julius and Collins, Katherine M and Juh{\'a}sz, Andr{\'a}s and Ruehle, Fabian and Welleck, Sean and Poesia, Gabriel and others},
  journal={arXiv preprint arXiv:2412.15184},
  year={2024}
}

@misc{claude2026riemann,
  author       = {{Claude}},
  title        = {More Than Two Thirds of the Zeros of the Riemann Zeta Function Are Simple and on the Critical Line},
  year         = {2026},
  howpublished = {Anthropic manuscript},
  url          = {https://www-cdn.anthropic.com/95c246936988e43127bc6b2ceb7077c1dad2d68e.pdf}
}

@misc{openai2026unitdistance,
  author       = {{OpenAI}},
  title        = {Planar Point Sets with Many Unit Distances},
  year         = {2026},
  howpublished = {Manuscript},
  url          = {https://cdn.openai.com/pdf/74c24085-19b0-4534-9c90-465b8e29ad73/unit-distance-proof.pdf}
}

@article{jang2025point,
  author        = {Uijeong Jang and Ernest K. Ryu},
  title         = {Point Convergence of Nesterov's Accelerated Gradient Method: An AI-Assisted Proof},
  year          = {2025},
  journal       = {arXiv preprint arXiv:2510.23513},
  eprint        = {2510.23513},
  archivePrefix = {arXiv},
  url           = {https://arxiv.org/abs/2510.23513}
}

@article{bubeck2025early,
  author        = {S{\'e}bastien Bubeck and Christian Coester and Ronen Eldan and Timothy Gowers and Yin Tat Lee and Alexandru Lupsasca and Mehtaab Sawhney and Robert Scherrer and Mark Sellke and Brian K. Spears and Derya Unutmaz and Kevin Weil and Steven Yin and Nikita Zhivotovskiy},
  title         = {Early Science Acceleration Experiments with {GPT-5}},
  year          = {2025},
  journal       = {arXiv preprint arXiv:2511.16072},
  eprint        = {2511.16072},
  archivePrefix = {arXiv},
  url           = {https://arxiv.org/abs/2511.16072}
}

@article{tao2026mathematics,
  title={Mathematics in the Age of AI},
  author={Tao, Terence},
  journal={arXiv preprint arXiv:2608.16753},
  year={2026}
}

@article{zhao2026opsd,
  title={Self-distilled reasoner: On-policy self-distillation for large language models},
  author={Zhao, Siyan and Xie, Zhihui and Liu, Mengchen and Huang, Jing and Pang, Guan and Chen, Feiyu and Grover, Aditya},
  journal={arXiv preprint arXiv:2601.18734},
  year={2026}
}

@article{dekoninck2026hmmt,
      title={Beyond Benchmarks: MathArena as an Evaluation Platform for Mathematics with LLMs}, 
      author={Jasper Dekoninck and Nikola Jovanović and Tim Gehrunger and Kári Rögnvaldsson and Ivo Petrov and Chenhao Sun and Martin Vechev},
      year={2026},
      eprint={2605.00674},
      archivePrefix={arXiv},
      primaryClass={cs.CL},
      url={https://arxiv.org/abs/2605.00674}, 
}

@inproceedings{gao2025omni,
  title={Omni-math: A universal olympiad level mathematic benchmark for large language models},
  author={Gao, Bofei and Song, Feifan and Yang, Zhe and Cai, Zefan and Miao, Yibo and Dong, Qingxiu and Li, Lei and Ma, Chenghao and Chen, Liang and Tang, Zhengyang and others},
  booktitle={International Conference on Learning Representations},
  volume={2025},
  pages={100540--100569},
  year={2025}
}

@article{shen2026purified,
  title={Purified OPSD: On-Policy Self-Distillation Without Losing How to Think},
  author={Shen, Zhanming and Tong, Jintao and Yan, Shaotian and Shen, Chen and Chen, Hao and Ye, Wentao and Hu, Xiaomeng and Miao, Rui and Wang, Haobo and Zhao, Junbo and others},
  journal={arXiv preprint arXiv:2607.02234},
  year={2026}
}

@article{he2026self,
  title={Self-distillation zero: Self-revision turns binary rewards into dense supervision},
  author={He, Yinghui and Kaur, Simran and Bhaskar, Adithya and Yang, Yongjin and Liu, Jiarui and Ri, Narutatsu and Fowl, Liam and Panigrahi, Abhishek and Chen, Danqi and Arora, Sanjeev},
  journal={arXiv preprint arXiv:2604.12002},
  year={2026}
}

@article{kaur2026rethinking,
  title={Rethinking On-Policy Self-Distillation for Thinking Models},
  author={Kaur, Simran and Ri, Narutatsu and He, Yinghui and Fowl, Liam and Arora, Sanjeev},
  journal={arXiv preprint arXiv:2607.05184},
  year={2026}
}

@article{penaloza2026privileged,
  title={Privileged information distillation for language models},
  author={Penaloza, Emiliano and Vattikonda, Dheeraj and Gontier, Nicolas and Lacoste, Alexandre and Charlin, Laurent and Caccia, Massimo},
  journal={arXiv preprint arXiv:2602.04942},
  year={2026}
}

@article{yang2026self,
  title={Self-distilled rlvr},
  author={Yang, Chenxu and Qin, Chuanyu and Si, Qingyi and Chen, Minghui and Gu, Naibin and Yao, Dingyu and Lin, Zheng and Wang, Weiping and Wang, Jiaqi and Duan, Nan},
  journal={arXiv preprint arXiv:2604.03128},
  year={2026}
}

@article{yu2026mismatch,
  title={Mismatch Matters: On-Policy Distillation Beyond Token Agreement},
  author={Yu, Zichao and Yu, Chengzhi and Xu, Shengze and Han, Yujin and Jiang, Bingqing and Wang, Xu and Zou, Difan},
  journal={arXiv preprint arXiv:2608.09836},
  year={2026}
}

@article{li2026demopsd,
  title={DemoPSD: Disagreement-Modulated Policy Self-Distillation},
  author={Li, Yunhe and Shi, Hao and Liu, Wenhao and Ruan, Mengzhe and Hou, Hanxu and Dai, Zhongxiang and Qiu, Shuang and Song, Linqi},
  journal={arXiv preprint arXiv:2607.02502},
  year={2026}
}

@article{peng2026diagnosing,
  title={Diagnosing and Mitigating Thinking Collapse in On-Policy Self-Distillation},
  author={Peng, Keqin and Li, Chen and Ouyang, Yuanxin and Yuan, Yancheng and Ding, Liang},
  journal={arXiv preprint arXiv:2607.10805},
  year={2026}
}

@article{xie2026trust,
  title={Trust Region Policy Distillation},
  author={Xie, Zhengpeng and Zhang, Li Lyna and Xie, Zeke and Yang, Mao},
  journal={arXiv preprint arXiv:2607.04751},
  year={2026}
}

@article{cobbe2021training,
  title={Training verifiers to solve math word problems},
  author={Cobbe, Karl and Kosaraju, Vineet and Bavarian, Mohammad and Chen, Mark and Jun, Heewoo and Kaiser, Lukasz and Plappert, Matthias and Tworek, Jerry and Hilton, Jacob and Nakano, Reiichiro and others},
  journal={arXiv preprint arXiv:2110.14168},
  year={2021}
}

@article{hendrycks2021measuring,
  title={Measuring mathematical problem solving with the math dataset},
  author={Hendrycks, Dan and Burns, Collin and Kadavath, Saurav and Arora, Akul and Basart, Steven and Tang, Eric and Song, Dawn and Steinhardt, Jacob},
  journal={arXiv preprint arXiv:2103.03874},
  year={2021}
}

@inproceedings{he2024olympiadbench,
  title={Olympiadbench: A challenging benchmark for promoting agi with olympiad-level bilingual multimodal scientific problems},
  author={He, Chaoqun and Luo, Renjie and Bai, Yuzhuo and Hu, Shengding and Thai, Zhen and Shen, Junhao and Hu, Jinyi and Han, Xu and Huang, Yujie and Zhang, Yuxiang and others},
  booktitle={Proceedings of the 62nd Annual Meeting of the Association for Computational Linguistics (Volume 1: Long Papers)},
  pages={3828--3850},
  year={2024}
}

@article{glazer2024frontiermath,
  title={Frontiermath: A benchmark for evaluating advanced mathematical reasoning in ai},
  author={Glazer, Elliot and Erdil, Ege and Besiroglu, Tamay and Chicharro, Diego and Chen, Evan and Gunning, Alex and Olsson, Caroline Falkman and Denain, Jean-Stanislas and Ho, Anson and Santos, Emily de Oliveira and others},
  journal={arXiv preprint arXiv:2411.04872},
  year={2024}
}

@inproceedings{hou-etal-2026-smart,
    title = "{SMART}: Evaluating {LLM}s' Mathematical Reasoning via a Human Cognitive Process-Inspired Benchmark",
    author = "Hou, Yujie  and
      Wang, Mei  and
      Zhong, Yaoyao  and
      Zhang, Ting  and
      Ma, Xuetao  and
      Huang, Hua",
    editor = "Liakata, Maria  and
      Moreira, Viviane P.  and
      Zhang, Jiajun  and
      Jurgens, David",
    booktitle = "Proceedings of the 64th Annual Meeting of the {A}ssociation for {C}omputational {L}inguistics (Volume 1: Long Papers)",
    month = jul,
    year = "2026",
    address = "San Diego, California, United States",
    publisher = "Association for Computational Linguistics",
    url = "https://aclanthology.org/2026.acl-long.1638/",
    doi = "10.18653/v1/2026.acl-long.1638",
    pages = "35426--35452",
    ISBN = "979-8-89176-390-6",
}

@article{zheng2026unmasking,
  title={Unmasking reasoning processes: A process-aware benchmark for evaluating structural mathematical reasoning in LLMs},
  author={Zheng, Xiang and Zhai, Weiqi and Wang, Wei and Yang, Boyu and Li, Wenbo and Luo, Ruixiang and Sun, Haoxiang and Wang, Yucheng and Li, Zhengze and Wang, Meng and others},
  journal={arXiv preprint arXiv:2602.00564},
  year={2026}
}

@article{li2026learning,
  title={Learning to Reason with Insight for Informal Theorem Proving},
  author={Li, Yunhe and Shi, Hao and Deng, Bowen and Wang, Wei and Ruan, Mengzhe and Hou, Hanxu and Dai, Zhongxiang and Gao, Siyang and Wang, Chao and Qiu, Shuang and others},
  journal={arXiv preprint arXiv:2604.16278},
  year={2026}
}

@inproceedings{amini2019mathqa,
  title={Mathqa: Towards interpretable math word problem solving with operation-based formalisms},
  author={Amini, Aida and Gabriel, Saadia and Lin, Shanchuan and Koncel-Kedziorski, Rik and Choi, Yejin and Hajishirzi, Hannaneh},
  booktitle={Proceedings of the 2019 Conference of the North American Chapter of the Association for Computational Linguistics: Human Language Technologies, Volume 1 (Long and Short Papers)},
  pages={2357--2367},
  year={2019}
}

@inproceedings{miao-etal-2020-diverse,
    title = "A Diverse Corpus for Evaluating and Developing {E}nglish Math Word Problem Solvers",
    author = "Miao, Shen-yun  and
      Liang, Chao-Chun  and
      Su, Keh-Yih",
    editor = "Jurafsky, Dan  and
      Chai, Joyce  and
      Schluter, Natalie  and
      Tetreault, Joel",
    booktitle = "Proceedings of the 58th Annual Meeting of the Association for Computational Linguistics",
    month = jul,
    year = "2020",
    address = "Online",
    publisher = "Association for Computational Linguistics",
    url = "https://aclanthology.org/2020.acl-main.92/",
    doi = "10.18653/v1/2020.acl-main.92",
    pages = "975--984",
}

@inproceedings{patel2021nlp,
  title={Are NLP models really able to solve simple math word problems?},
  author={Patel, Arkil and Bhattamishra, Satwik and Goyal, Navin},
  booktitle={Proceedings of the 2021 conference of the North American chapter of the association for computational linguistics: human language technologies},
  pages={2080--2094},
  year={2021}
}

@inproceedings{chen2023theoremqa,
  title={Theoremqa: A theorem-driven question answering dataset},
  author={Chen, Wenhu and Yin, Ming and Ku, Max and Lu, Pan and Wan, Yixin and Ma, Xueguang and Xu, Jianyu and Wang, Xinyi and Xia, Tony},
  booktitle={Proceedings of the 2023 Conference on Empirical Methods in Natural Language Processing},
  pages={7889--7901},
  year={2023}
}

@inproceedings{liu2024mathbench,
  title={Mathbench: Evaluating the theory and application proficiency of llms with a hierarchical mathematics benchmark},
  author={Liu, Hongwei and Zheng, Zilong and Qiao, Yuxuan and Duan, Haodong and Fei, Zhiwei and Zhou, Fengzhe and Zhang, Wenwei and Zhang, Songyang and Lin, Dahua and Chen, Kai},
  booktitle={Findings of the Association for Computational Linguistics: ACL 2024},
  pages={6884--6915},
  year={2024}
}

@article{khan2025executable,
  title={Executable Functional Abstractions: Inferring Generative Programs for Advanced Math Problems},
  author={Khan, Zaid and Stengel-Eskin, Elias and Prasad, Archiki and Cho, Jaemin and Bansal, Mohit},
  journal={arXiv preprint arXiv:2504.09763},
  year={2025}
}

@article{nguyen2026avsd,
  title={AVSD: Adaptive-View Self-Distillation by Balancing Consensus and Teacher-Specific Privileged Signals},
  author={Nguyen, Duy and Xiao, Hanqi and Prasad, Archiki and Khan, Zaid and Das, Anirban and Zhang, Austin and Sahu, Sambit and Lee, Hyunji and Stengel-Eskin, Elias and Bansal, Mohit},
  journal={arXiv preprint arXiv:2605.20643},
  year={2026}
}
\bibliographystyle{iclr2027_conference}

\newpage
\appendix
\section{Data Curation Details}
\label{app:data}
\subsection{Dataset Filtering and Correction}
\label{app:data-flilter}

To reduce annotation noise in the original HLE, we draw from the community-verified \textsc{HLE-Verified} release~\citep{zhai2026hle-verified}, retaining only its \emph{Gold} and \emph{Revision} subsets. We randomly sample $200$ text-only problems and conduct an additional expert review for primitive-level evaluation. This process excludes $18$ unsuitable problems due to ill-posed or ambiguous questions, incorrect or unverifiable answers, non-self-contained tasks, insufficient rationales, exhaustive-computation requirements, or other quality issues. For $17$ of these $18$ problems, no valid gold primitive could be established, yielding a final benchmark of $182$ problems.

Expert review further identifies residual errors in the official answers of $15$ retained problems, including $13$ mathematical errors and $2$ corrupted answer strings, which we correct in our benchmark. Re-evaluating the strongest model on these corrected answers flips $11$ predictions from incorrect to correct, with no reverse flips, providing an additional consistency check on the revisions.

Because gold primitives are derived with access to reference solutions, we also audit them for answer leakage. We search the concatenated primitive fields for bounded occurrences of the gold answer, using token boundaries designed to avoid spurious numeral matches. This audit initially identifies $4$ affected primitives. We manually revise them to remove the leaked answers while preserving their structural content, and verify that no direct answer-string leakage remains in the final benchmark.

\subsection{Primitive Type and Family}
\label{app:primitive-family}

To make the definition operational and to structure evaluation, we anchor every primitive to a frozen taxonomy of ten primitive \emph{types} (plus a \texttt{computation\_only} escape hatch for problems that genuinely have no kernel primitive in the sense of Definition~\ref{def:primitive}), as presented in Table~\ref{tab:primitive-types}.

\begin{table}[htbp]
\centering
\footnotesize
\begin{tabular}{@{}l p{8cm}@{}}
\toprule
\textbf{Type} & \textbf{The essential property is\ldots} \\
\midrule
\texttt{reformulation} & a change of language/encoding/viewpoint makes the problem tractable \\
\texttt{hidden\_structure} & the objects secretly carry a known structure (group, graph, metric space, \ldots) \\
\texttt{invariant\_or\_monotonicity} & a preserved or monotone quantity constrains all transformations \\
\texttt{theorem\_applicability} & the task reduces to verifying the hypotheses of a known theorem \\
\texttt{duality\_or\_representation} & a dual / spectral / transformed viewpoint simplifies the problem \\
\texttt{extremal\_principle} & a maximal/minimal/extreme object forces structure \\
\texttt{construction\_or\_witness} & an explicit object/certificate/counterexample is built \\
\texttt{reduction\_or\_embedding} & the problem is reduced/embedded into a known canonical one \\
\texttt{obstruction\_or\_contradiction} & the opposite assumption violates a necessary constraint \\
\texttt{induction\_or\_recursion} & a recursive/self-similar structure with a strengthened hypothesis \\
\bottomrule
\end{tabular}
\caption{The ten primitive types.}
\label{tab:primitive-types}
\end{table}

\begin{table}[htbp]\centering\footnotesize
\setlength{\tabcolsep}{4pt}\renewcommand{\arraystretch}{1.25}
\begin{tabular}{@{}l p{5.2cm} p{3cm} p{3cm}@{}}
\toprule
Family & Routed primitive types & Gate (\textsc{what}) & Mechanism (\textsc{how}) \\
\midrule
\textsc{Recast}
& \texttt{reformulation}, \texttt{hidden\_structure}, \texttt{reduction\_or\_embedding}, \texttt{duality\_or\_representation}, \texttt{theorem\_applicability}
& target setting $Y$ (equivalent form / hidden structure / canonical target / dual / named theorem)
& how $Y$ unlocks it; for theorems, hypotheses verified \\
\addlinespace
\textsc{Witness}
& \texttt{construction\_or\_witness}, \texttt{extremal\_principle}, \texttt{invariant\_or\_monotonicity}
& the object (construction / certificate / extremal / invariant--monotone)
& its decisive property: built to satisfy / extremality / invariance forces it \\
\addlinespace
\textsc{Argument}
& \texttt{obstruction\_or\_contradiction}, \texttt{induction\_or\_recursion}
& the load-bearing claim (specific obstruction / strengthened hypothesis)
& why the opposite fails / how the recursion closes \\
\bottomrule
\end{tabular}
\caption{The primitive families. Each of the ten frozen primitive types routes to exactly one family.}
\label{tab:primitive-families}
\end{table}

Further, since the same problem can often be solved through conceptually different routes, we further assign the primary primitive type as presented in Table~\ref{tab:primitive-families} of each problem to one of three \emph{families}: \textsc{Recast}, \textsc{Witness}, or \textsc{Argument}, as summarized below:
\begin{itemize}[leftmargin=1em]
\item \textbf{\textsc{Recast}}: The load-bearing object is the \emph{target setting} $Y$ into which the original problem is moved, such as an equivalent reformulation, a hidden known structure, a canonical target problem, a dual or representational viewpoint, or a named theorem. The gate asks whether the model recasts the problem into the same $Y$, or a provably equivalent one up to trivial restatement. The mechanism asks how this target setting unlocks the solution, such as the reduction step actually used, or, in theorem-based cases, whether the theorem's hypotheses are correctly verified. Moving to a different setting, naming the wrong theorem, or remaining entirely within the original framing constitutes a gate mismatch; identifying a coarser or weaker target receives partial credit.

\item \textbf{\textsc{Witness}}: The load-bearing object is a specific mathematical \emph{object}: a construction, certificate, witness, extremal object, invariant, or monotone quantity. The gate asks whether the model identifies the same object, or an equivalent object that plays the identical role in the solution. The mechanism asks for its decisive property: for example, why the construction satisfies the required constraint, why extremality forces the conclusion, or why invariance forbids or forces the desired outcome. An object of the correct general kind but without the decisive feature receives partial credit, such as a near-construction that fails the needed property or a monotone quantity that is not the one driving the proof.

\item \textbf{\textsc{Argument}}: The load-bearing object is the central \emph{claim} on which the proof depends, such as the obstruction contradicted by the opposite assumption or the strengthened induction hypothesis needed for the recursive step. The gate asks whether the model's reasoning hinges on the same claim; merely naming the proof style, such as ``by contradiction'' or ``by induction'', is not sufficient. The mechanism asks for the logical step itself, such as why the opposite assumption violates the constraint or how the strengthened hypothesis closes the induction. Stating only the proof style without the specific load-bearing claim is a gate mismatch, while a related but loosely stated or under-strengthened claim receives partial credit.

\end{itemize}

\subsection{Primitive Scoring Protocol}
\label{app:annnotate}
Mathematical primitive correctness is evaluated against the annotated \emph{gold primitive}, with the problem's full reference solution (\emph{rationale}) supplied as supporting context. The rationale is provided only so the judge can recognize coarser phrasings, equivalent paraphrases, or alternative routes that the reference solution shows to be valid, i.e.\ so a correct primitive is not penalized merely for failing to match the terse gold annotation verbatim, while never requiring the model to reproduce the downstream execution. 

All three primitive families share the same three-verdict decomposition $(V,\sigma_{\text{gate}},\sigma_{\text{mech}})$ and the deterministic scoring rule in \S~\ref{subsec:primitive-bench}. Here, $V\in\{0,1\}$ is a binary validity indicator (is the output a genuine primitive rather than a generic remark, a bare technique name, or pure computation), while the two graded slots $\sigma_{\text{gate}},\sigma_{\text{mech}}\in\{0,\tfrac12,1\}$ evaluate whether the response identifies the essential \textsc{what} (the \emph{gate}) and explains the operative \textsc{how} (the \emph{mechanism}). The difference across families is the type of mathematical object over which these two slots are instantiated, ensuring that ``the same idea'' is matched against the appropriate target. Because $V$ is binary and $\sigma_{\text{gate}},\sigma_{\text{mech}}\in
\{0,\tfrac12,1\}$, the score lies in $[0,1]$ but is restricted to the finite set, and we have $\mathrm{Score}\in\{0,\;0.30,\;0.40,\;0.50,\;0.60,\;0.80,\;1.00\}.$

All primitive judgments are produced by GPT-5.4-High at temperature 0. To validate the judge, we sampled $200$ (model, problem) pairs, $100$ judged primitive-correct and $100$ judged primitive-incorrect, stratified across the $12$ evaluated models. Three annotators with graduate-level mathematical training, blind to the judge's verdict, independently labeled $\mathrm{PrimitiveAcc}$. Judge--human agreement is $95.5\%$ (Cohen's $\kappa=0.91$), with $9$ disagreements.

Because each problem is scored against a single gold primitive, a response following a genuinely different but valid conceptual route may be under-credited. To assess this limitation, we further examined the $100$ primitive-incorrect cases in the same validation sample ($8$--$9$ per model, fixed seed): exactly $1$ case ($1\%$) contained a valid alternative primitive missed by the single-gold rubric. We additionally audited all $32$ problems marked primitive-incorrect for the strongest model (gpt-5.4, $\mathrm{Score}<0.8$), identifying exactly $1$ such case. These audits suggest that, although the single-gold rubric may conservatively underestimate valid primitive generation, under-counting due to alternative conceptual routes is rare in the audited cases.

\subsection{Illustrative Example of Mathematical Primitives}
To make the notion of a mathematical primitive concrete, we provide a representative example in Figure~\ref{fig:primitive_example}. The full rationale contains the complete derivation needed to solve the problem, whereas the primitive isolates the underlying structural idea that makes the derivation possible. Due to space constraints, we show only the \texttt{core\_concept} field in the figure. This example illustrates that a primitive is not a shortened proof, but a compact representation of the key mathematical structure underlying the solution.

\begin{figure}[h]
    \centering
    \begin{boxK}[colback=gray!3!white,colframe=black!45!gray,fontupper=\scriptsize]

        \begingroup
        \setlength{\abovedisplayskip}{3pt}
        \setlength{\belowdisplayskip}{3pt}
        \setlength{\abovedisplayshortskip}{3pt}
        \setlength{\belowdisplayshortskip}{3pt}

        {\fontsize{9}{10.5}\selectfont\bfseries Question}
        \par\vspace{2pt}

        Let
        \[
        L:=\mathbb{Q}\left(
        \sqrt{(2+\sqrt{2})(3+\sqrt{3})},
        \sqrt{2},
        \sqrt{3}
        \right).
        \]
        What is the Galois group of $L/\mathbb{Q}$?

        \par\vspace{4pt}
        {\color{black!25}\hrule height 0.4pt}
        \vspace{5pt}

        {\fontsize{9}{10.5}\selectfont\bfseries Rationale}
        \par\vspace{3pt}

        Set $\alpha=\sqrt2$, $\beta=\sqrt3$, and
        \[
        \gamma=\sqrt{(2+\alpha)(3+\beta)}.
        \]
        Let $K=\mathbb{Q}(\alpha,\beta)$. Then $K/\mathbb{Q}$ is biquadratic, hence Galois with $[K:\mathbb{Q}]=4$ and $\mathrm{Gal}(K/\mathbb{Q})\cong C_2\times C_2$.

        \smallskip
        \textbf{Step 1: $[L:K]=2$ and $[L:\mathbb{Q}]=8$}

        We have $\gamma^2=(2+\alpha)(3+\beta)\in K$, so $L=K(\gamma)$ and $[L:K]\le 2$.
        To see $\gamma\notin K$, it suffices to show $a:=(2+\alpha)(3+\beta)$ is not a square in $K$.

        Consider the relative norm $N_{K/\mathbb{Q}(\alpha)}$. The nontrivial automorphism of $K/\mathbb{Q}(\alpha)$ sends $\beta\mapsto -\beta$, hence
        \[
        N_{K/\mathbb{Q}(\alpha)}(a)=a\cdot a^{\tau}
        =(2+\alpha)(3+\beta)\cdot(2+\alpha)(3-\beta)
        =(2+\alpha)^2(9-\beta^2)
        =6(2+\alpha)^2.
        \]
        If $a$ were a square in $K$, then its norm down to $\mathbb{Q}(\alpha)$ would be a square in $\mathbb{Q}(\alpha)$. But $6(2+\alpha)^2$ is a square in $\mathbb{Q}(\alpha)$ if and only if $6$ is a square in $\mathbb{Q}(\alpha)$. Now $6$ is not a square in $\mathbb{Q}(\sqrt2)$: if $(u+v\alpha)^2=6$ with $u,v\in\mathbb{Q}$, then the $\alpha$-coefficient gives $2uv=0$, so either $v=0$ (forcing $u^2=6$, impossible) or $u=0$ (forcing $2v^2=6$, i.e. $v^2=3$, impossible in $\mathbb{Q}$).

        Hence $a$ is not a square in $K$, so $[L:K]=2$ and therefore $[L:\mathbb{Q}]=8$.

        \smallskip
        \textbf{Step 2: Explicit automorphisms and normality}

        Define the involution
        \[
        \rho:\ \alpha\mapsto\alpha,\ \beta\mapsto\beta,\ \gamma\mapsto-\gamma.
        \]
        This generates $\mathrm{Gal}(L/K)\cong C_2$.

        Next extend the two generators of $\mathrm{Gal}(K/\mathbb{Q})$. Let $\sigma_0$ be the automorphism of $K$ with $\alpha\mapsto-\alpha,\ \beta\mapsto\beta$. Note that
        \[
        \sigma_0(\gamma^2)=(2-\alpha)(3+\beta).
        \]
        But
        \[
        \frac{(2-\alpha)(3+\beta)}{(2+\alpha)(3+\beta)}
        =\frac{2-\alpha}{2+\alpha}
        =3-2\alpha
        =(\alpha-1)^2,
        \]
        so $\sigma_0(\gamma^2)=(\alpha-1)^2\gamma^2$. Hence we can define an extension $\sigma\in\mathrm{Aut}(L)$ by
        \[
        \sigma:\ \alpha\mapsto-\alpha,\ \beta\mapsto\beta,\ 
        \gamma\mapsto(\alpha-1)\gamma,
        \]
        since then $\sigma(\gamma)^2=(\alpha-1)^2\gamma^2=\sigma_0(\gamma^2)$.

        Similarly let $\tau_0$ be the automorphism of $K$ with $\alpha\mapsto\alpha,\ \beta\mapsto-\beta$. Then
        \begin{align*}
            \tau_0(\gamma^2)=(2+\alpha)(3-\beta), \quad \text{and} \quad \frac{(2+\alpha)(3-\beta)}{(2+\alpha)(3+\beta)}
        =\frac{3-\beta}{3+\beta}
        =2-\beta
        =\left(\frac{\beta-1}{\alpha}\right)^2
        \end{align*}
        (using $(\beta-1)^2=4-2\beta$ and $\alpha^2=2$). Thus $\tau_0(\gamma^2)=\left(\frac{\beta-1}{\alpha}\right)^2\gamma^2$, and we define $\tau\in\mathrm{Aut}(L)$ by
        \[
        \tau:\ \alpha\mapsto\alpha,\ \beta\mapsto-\beta,\ 
        \gamma\mapsto\frac{\beta-1}{\alpha}\,\gamma.
        \]
        These maps show every sign-change on $\alpha,\beta$ extends to $L$, so $L/\mathbb{Q}$ is Galois. Since $[L:\mathbb{Q}]=8$, we now have $|\mathrm{Gal}(L/\mathbb{Q})|=8$.

        \smallskip
        \textbf{Step 3: Group structure}

        Compute squares on $\gamma$:
        \[
        \sigma^2(\gamma)
        =\sigma\big((\alpha-1)\gamma\big)
        =\sigma(\alpha-1)\,\sigma(\gamma)
        =(-\alpha-1)(\alpha-1)\gamma
        =-(\alpha^2-1)\gamma
        =-\gamma,
        \]
        so $\sigma^2=\rho$. Likewise,
        \[
        \tau^2(\gamma)
        =\tau\!\left(\frac{\beta-1}{\alpha}\gamma\right)
        =\tau\!\left(\frac{\beta-1}{\alpha}\right)\tau(\gamma)
        =\left(\frac{-\beta-1}{\alpha}\right)
        \left(\frac{\beta-1}{\alpha}\right)\gamma
        =-\frac{\beta^2-1}{\alpha^2}\gamma
        =-\frac{2}{2}\gamma
        =-\gamma,
        \]
        so $\tau^2=\rho$. In particular, $\sigma$ and $\tau$ have order $4$, and $\rho$ is central of order $2$.

        Finally compare $\sigma\tau$ and $\tau\sigma$ on $\gamma$. Let $u=\alpha-1$ and $v=(\beta-1)/\alpha$. Then $\sigma(v)=-v$ (because $\sigma(\alpha)=-\alpha$ and $\sigma(\beta)=\beta$), while $\tau(u)=u$. Hence
        \begin{align*}
            \sigma\tau(\gamma)
        =\sigma(v\gamma)
        =\sigma(v)\sigma(\gamma)
        =(-v)\,u\gamma
        =-(uv)\gamma, \quad \text{but} \quad \tau\sigma(\gamma)
        =\tau(u\gamma)
        =\tau(u)\tau(\gamma)
        =u\,v\gamma
        =(uv)\gamma.
        \end{align*}
        Therefore $\sigma\tau(\gamma)=-\tau\sigma(\gamma)$, i.e. $\sigma\tau=\rho\,\tau\sigma$. So the group is generated by $\sigma,\tau$ with relations
        \[
        \sigma^2=\tau^2=\rho,\quad
        \rho^2=1,\quad
        \rho\ \text{central},\quad
        \sigma\tau=\rho\,\tau\sigma,
        \]
        which is exactly the quaternion group of order $8$. Equivalently, every nontrivial element other than $\rho$ has order $4$, so the group cannot be dihedral; it is $Q_8$.

        \par\vspace{4pt}
        {\color{black!25}\hrule height 0.4pt}
        \vspace{5pt}

        {\fontsize{9}{10.5}\selectfont\bfseries Primitive}
        \par\vspace{3pt}

        \textbf{core\_concept.}
        The key insight is to track the added radical by its square class over the biquadratic base; preservation of that square class gives the lifts, and the orders of those lifts together with whether they commute are what pin down the isomorphism type.

        \endgroup
    \end{boxK}

    \vspace{-0.1in}
    \caption{An example illustrating the distinction between a full solution rationale and the corresponding mathematical primitive.}
    \label{fig:primitive_example}
\end{figure}

\subsection{Prompt on Extracting a Mathematical Primitive and Primitive Scoring}

During data curation, we use a teacher model to extract a mathematical primitive for each problem. As shown in Figure~\ref{fig:primitive_curation_prompt}, the teacher is asked to identify the essential mathematical property underlying the problem and explain the solution principle enabled by that property, while explicitly avoiding calculations or solution steps. We additionally require a structured primitive classification, including a primary type, optional secondary types, a classification rationale, and a confidence score.

We evaluate each predicted primitive using three complementary signals. First, the validity term $V$ checks whether the prediction contains a substantive mathematical primitive rather than a vacuous response, such as generic advice, a bare technique name, or computation without a conceptual claim (Figure~\ref{fig:primitive_validity_prompt}). For predictions that pass this check, we separately assess the correctness of the GATE and MECHANISM components using the reference solution and gold primitive (Figure~\ref{fig:primitive_correctness_prompt}). The GATE score $\sigma_{\mathrm{gate}}$ measures whether the prediction identifies the essential problem-specific idea, the ``what'' that unlocks the problem, while the MECHANISM score $\sigma_{\mathrm{mech}}$ measures whether it correctly explains the high-level ``how,'' namely why that idea enables a valid solution principle. Both correctness judgments emphasize semantic agreement rather than exact reproduction and do not penalize missing downstream execution details.

\begin{figure}[h]
    \centering
    \begin{boxK}[colback=royalblue!5!white, colframe=royalblue!75!black, fontupper=\ttfamily\tiny]
        \ You are given a language-only mathematics problem, its answer, and its rationale. Your task is to extract and classify the mathematical PRIMITIVE behind the rationale.
        \\
        \textbf{\#\# Definition:} \\
        A mathematical primitive is the essential mathematical property that makes the problem solvable, together with an explanation of how this property connects the problem to a valid solution principle. \\
        The primitive is not merely a theorem name, a technique, or a proof step. For example, ``use induction,'' ``apply Riesz representation,'' or ``use generating functions'' is insufficient unless it explains what property of the problem makes that method applicable.
        \\
        \textbf{\#\# Steps:} \\
        \textbf{Step 1: Identify the essential mathematical property.} \\
        Ask: What hidden structure, invariant, relation, decomposition, obstruction, or theorem condition makes the problem tractable? \\
        \textbf{Step 2: Identify the solution principle.} \\
        Ask: How does that property enable a valid theorem, reduction, construction, reformulation, contradiction, or proof strategy? Name the essential conceptual move(s) --- one OR several distinct key ideas, whichever the problem genuinely needs. \\
        \textbf{Step 3: Write the mathematical primitive.} \\
        Write 1--3 sentences combining the essential property and its connection to the solution principle.
        \\
        \textbf{\#\# CRITICAL CONSTRAINTS:}
        \begin{itemize}[leftmargin=0.15in,nosep]
            \item Strictly conceptual: no algebraic manipulation, equations, or step-by-step proof.
            \item HIGH-LEVEL but COMPLETE: state EVERY essential CONCEPTUAL idea the solution rests on (a problem may need ONE or SEVERAL distinct key observations --- include each one). Do NOT write out calculations, derivations, or intermediate numeric results.
        \end{itemize}

        \textbf{Step 4: Classify the primitive.} \\
        Choose exactly one primary primitive type and zero to three secondary primitive types. \\
        \textbf{CRITICAL CONSTRAINT:} You MUST use the exact string keys provided in the taxonomy below (e.g., ``reformulation'', ``hidden\_structure''). Do NOT modify, capitalize, or invent new types. \\
        The primary primitive type should describe the main mathematical mechanism that unlocks the solution. \\
        The secondary primitive types should describe additional mathematical ideas that support, enable, or refine the primary primitive. \\
        Do not classify based only on surface-level technique names. Classify based on the essential mathematical property and how it connects to the solution principle.
        \\
        \textbf{\#\# Primitive Taxonomy:}
        \begin{itemize}[leftmargin=0.15in,nosep]
            \item \textbf{reformulation}: The problem becomes tractable by rewriting it in a more useful mathematical language, representation, or equivalent form.

            \item \textbf{hidden\_structure}: The key is recognizing that the objects secretly have a known mathematical structure, such as a group, vector space, Hilbert space, graph, lattice, convex set, metric space, probability space, or algebraic structure.

            \item \textbf{invariant\_or\_monotonicity}: The key is identifying a quantity or property that is preserved, monotone, or constrains all valid transformations.

            \item \textbf{theorem\_applicability}: The key is recognizing that the problem reduces to verifying the assumptions of a known theorem.

            \item \textbf{duality\_or\_representation}: The key is passing to a dual, transformed, spectral, functional, or representational viewpoint where the problem becomes simpler.

            \item \textbf{extremal\_principle}: The key is choosing a maximal, minimal, largest, smallest, boundary, or otherwise extreme object and exploiting its extremality.

            \item \textbf{construction\_or\_witness}: The key is explicitly constructing an object, example, counterexample, certificate, function, sequence, graph, or witness.

            \item \textbf{reduction\_or\_embedding}: The key is reducing the problem to another known or canonical problem, or embedding it into a setting where a known result applies.

            \item \textbf{obstruction\_or\_contradiction}: The key is identifying a necessary structural constraint that the opposite assumption would violate.

            \item \textbf{induction\_or\_recursion}: The key is recognizing a recursive, self-similar, or inductive structure and choosing an induction hypothesis that preserves the essential property.
        \end{itemize}

        \textbf{\#\# DO NOT SOLVE OR SUMMARIZE THE SOLUTION:}
        \begin{itemize}[leftmargin=0.15in,nosep]
            \item Do NOT state or compute the final answer; no specific numeric result, final value, or boxed expression.
            \item Do NOT write a full derivation or proof.
            \item Do NOT write out the calculations or the step-by-step derivation. State all the essential conceptual primitive(s) the solution rests on --- including each distinct key idea it genuinely needs --- but not the execution that follows.
        \end{itemize}

        Extract and classify the mathematical primitive behind the rationale.
        \\
        \textbf{\#\# Output Format:} \\
        Output ONLY a valid JSON object in the exact format below, with no markdown code blocks formatting (e.g., do not wrap in ` ` ` json), just the raw JSON: \\
        \{ \\
        \ \ \ \ "essential\_property": "...", \\
        \ \ \ \ "solution\_principle": "...", \\
        \ \ \ \ "core\_concept": "...", \\
        \ \ \ \ "primary\_primitive\_type": "...", \\
        \ \ \ \ "secondary\_primitive\_types": ["..."], \\
        \ \ \ \ "classification\_rationale": "...", \\
        \ \ \ \ "confidence": <integer between 1 and 5> \\
        \}
        \\
        \textbf{\#\# Problem:} \\
        {[}PROBLEM{]}
        \\
        \textbf{\#\# Answer:} \\
        {[}GOLD\_ANSWER{]}
        \\
        \textbf{\#\# Rationale:}\\
        {[}GOLD\_RATIONALE{]}
    \end{boxK}
    \vspace{-0.1in}
    \caption{Prompt used during data curation to generate mathematical primitives and their structured annotations.}
    \label{fig:primitive_curation_prompt}
\end{figure}

\begin{figure}[h]
    \centering
    \begin{boxK}[colback=royalblue!5!white, colframe=royalblue!75!black, fontupper=\ttfamily\tiny]
        \ You are checking whether a candidate mathematical ``primitive'' for a mathematics problem is VACUOUS --- i.e., not really a primitive at all. Look only at the candidate below.
        \\
        A mathematical primitive names the essential property that makes the problem solvable and explains how it connects to a solution principle.
        \\
        Mark verdict = ``invalid'' ONLY if the candidate is VACUOUS, i.e., ANY of:
        \begin{itemize}[leftmargin=0.15in]
            \item \textbf{``generic''}: vague advice with no problem-specific content (``use a clever substitution'', ``find the pattern'').
            \item \textbf{``technique\_name\_only''}: names a method/theorem but never states the property of THIS problem that makes it apply.
            \item \textbf{``computation\_only''}: only arithmetic/numeric work, with no conceptual claim.
        \end{itemize}

        Otherwise verdict = ``valid'', failure = ``none''.
        \\
        Do NOT mark invalid for stating a conclusion or a final result, and do NOT mark invalid for being a full derivation --- over-completeness is NOT vacuity. A thorough or conclusion-stating answer can still contain a valid primitive.
        \\
        \textbf{\#\# Candidate Primitive:} \\
        essential\_property: [MODEL essential\_property] \\
        solution\_principle: [MODEL solution\_principle] \\
        core\_concept: [MODEL core\_concept]
        \\
        \textbf{\#\# Output Format:} \\
        Return JSON only: \\
        \{``verdict'': ``valid''|``invalid'', ``failure'': ``none''|``generic''|``technique\_name\_only''|``computation\_only'', ``evidence'': ``<one short sentence>''\}
    \end{boxK}
    \vspace{-0.1in}
    \caption{Prompt for assessing primitive validity $V$ by detecting vacuous predictions.}
    \label{fig:primitive_validity_prompt}
\end{figure}

\begin{figure}[h]
    \centering
    \begin{boxK}[colback=royalblue!5!white, colframe=royalblue!75!black, fontupper=\ttfamily\tiny]
        \ You are judging whether a MODEL's candidate mathematical primitive for a math problem is CORRECT on ONE dimension, using a known-correct REFERENCE SOLUTION. Judge correctness FOR THIS PROBLEM --- not reproduction of any particular wording.
        \\
        \textbf{\#\# GROUND RULES (read carefully):}
        \begin{itemize}[leftmargin=0.15in]
            \item The REFERENCE SOLUTION below is correct and complete. The GOLD PRIMITIVE is a terse distillation of its key idea, written WITH the full solution in hand. The MODEL only saw the problem.
            \item Judge whether the MODEL's primitive correctly captures the [DIM] that unlocks this problem, as evidenced by the reference solution.
            \item CREDIT a correct primitive even if it is (i) phrased more coarsely or less precisely than the gold primitive, or (ii) a different but valid route that the reference solution shows would work.
            \item Do NOT require the model to reproduce the gold primitive's exact framing, or to name the same theorem if its statement is equivalent.
            \item Do NOT re-solve from scratch; judge the model's primitive against the reference solution.
            \item Do NOT consider whether the model reached any final answer --- judge the primitive content only.
            \item A primitive here is the SINGLE CORE high-level idea (the dominant ``aha''). Everything DOWNSTREAM of it is execution, NOT part of the primitive, and its absence must NOT lower the score: secondary lemmas, second-step ideas, lower-bound / sharpness / optimality arguments, exact constants, specific numeric values, and calculation steps are what the core primitive ENABLES. Judge only whether the model has the correct CORE idea. (A vague, non-problem-specific statement, however, is still not a primitive.)
        \end{itemize}

        The dimension you are judging is the \textbf{[DIM]}: \\
        {[}DIM\_DEF{]}
        \\
        \textbf{\#\# Scoring (CORRECTNESS, not reproduction):}
        \begin{itemize}[leftmargin=0.15in]
            \item \textbf{``match''}: [C\_MATCH]
            \item \textbf{``partial''}: [C\_PARTIAL]
            \item \textbf{``mismatch''}: [C\_MISMATCH]
        \end{itemize}

        \textbf{\#\# Problem:} \\
        {[}PROBLEM{]}
        \\
        \textbf{\#\# Reference Solution (correct):} \\
        {[}REFERENCE SOLUTION{]}
        \\
        \textbf{\#\# Gold Primitive (terse pointer to the essential idea):} \\
        essential\_property: [GOLD essential\_property] \\
        solution\_principle: [GOLD solution\_principle] \\
        core\_concept: [GOLD core\_concept]
        \\
        \textbf{\#\# Model Primitive (judge this):} \\
        essential\_property: [MODEL essential\_property] \\
        solution\_principle: [MODEL solution\_principle] \\
        core\_concept: [MODEL core\_concept]
        \\
        \textbf{\#\# Output Format:} \\
        Return JSON only: \\
        \{``agreement'': ``match''|``partial''|``mismatch'', ``evidence'': ``<one short sentence>''\}
    \end{boxK}
    \vspace{-0.1in}
    \caption{Prompt for assessing primitive correctness along the GATE and MECHANISM dimensions, corresponding to $\sigma_{\mathrm{gate}}$ and $\sigma_{\mathrm{mech}}$.}
    \label{fig:primitive_correctness_prompt}
\end{figure}

\section{Additional Details on Diagnostic Evaluation}
\label{app:diagno-detail}

\subsection{Comparison with Full-HLE Performance}
\label{app:performance}
Notably, the absolute answer accuracies in Table~\ref{tab:gap-family} are substantially higher than the commonly reported performance of these model families on the full HLE benchmark. This difference should not be interpreted as a contradiction, for two reasons: \ding{182} \hlei{} is a curated scorable subset rather than the full HLE distribution; and \ding{183} during benchmark construction, human experts removed examples with annotation noise or evaluation ambiguity in the original HLE~\citep{phan2025lastexam} and \textsc{HLE-Verified}~\citep{zhai2026hle-verified} sources, including fundamentally incorrect reference answers, logical gaps in rationales, and irregular formatting that hinders automated evaluation. 

\subsection{Prompt on Generating the Primitive and Plan using GPT-5.4}

We use GPT-5.4 to generate two forms of teacher guidance for the Execution analysis: a Teacher Primitive, which captures the load-bearing mathematical structure following Definition~\ref{def:primitive}, and a Teacher Plan, which provides a step-by-step solution outline. The prompts used to generate these two forms of guidance are provided below in Figures \ref{fig:teacher_primitive_prompt} and \ref{fig:teacher_plan_prompt}.

\begin{figure}[h]
    \centering
    \begin{boxK}[colback=royalblue!5!white, colframe=royalblue!75!black, fontupper=\ttfamily\tiny]
    \ You are given a language-only mathematics problem --- the problem ONLY, with no solution. Your task is to produce the single kernel mathematical PRIMITIVE needed to solve it: the essential structural understanding that makes the problem tractable. Do NOT solve the problem.
        \\
        \textbf{\#\# Definition:} \\
        A mathematical primitive is the essential mathematical structure or property that makes the problem solvable, together with an explanation of how this structure connects the problem to a valid solution principle. \\
        The primitive is not merely a theorem name, a technique, or a proof step. For example, ``use induction,'' ``apply Riesz representation,'' or ``use generating functions'' is insufficient unless it explains what property of the problem makes that method applicable.
        \\
        \textbf{\#\# Steps:} \\
        \textbf{Step 1: Identify the essential mathematical property.} \\
        Ask: What hidden structure, invariant, relation, decomposition, obstruction, or theorem condition makes the problem tractable? \\
        \textbf{Step 2: Identify the solution principle.} \\
        Ask: How does that property enable a valid theorem, reduction, construction, reformulation, contradiction, or proof strategy? Name the essential conceptual move(s) --- one OR several distinct key ideas, whichever the problem genuinely needs. \\
        \textbf{Step 3: Write the mathematical primitive.} \\
        Write 1--3 sentences capturing the single high-level ``Aha!'' --- the central structural understanding that makes the problem tractable.
        \\
        \textbf{\#\# CRITICAL CONSTRAINTS:}
        \begin{itemize}[leftmargin=0.15in,nosep]
            \item Strictly conceptual: no algebraic manipulation, equations, or step-by-step proof.
            \item HIGH-LEVEL but COMPLETE: state EVERY essential CONCEPTUAL idea the solution rests on (a problem may need ONE or SEVERAL distinct key observations --- include each one). Do NOT write out calculations, derivations, or intermediate numeric results.
            \item Contrast: ``recognize the configuration is secretly a Cayley graph'' is a mathematical primitive; ``build the Cayley graph, compute its spectrum, then sum the eigenvalues to get the count'' is a solution outline --- produce the former, never the latter.
        \end{itemize}

        \textbf{\#\# DO NOT SOLVE OR SUMMARIZE THE SOLUTION:}
        \begin{itemize}[leftmargin=0.15in,nosep]
            \item Do NOT state or compute the final answer; no specific numeric result, final value, or boxed expression.
            \item Do NOT write a full derivation or proof.
            \item Do NOT write out the calculations or the step-by-step derivation. State all the essential conceptual primitive(s) the solution rests on --- including each distinct key idea it genuinely needs --- but not the execution that follows.
        \end{itemize}
        Produce the essential mathematical primitive(s): the key thing(s) to notice before starting to solve.
        \\
        \textbf{\#\# Output Format:} \\
        Output ONLY a valid JSON object in the exact format below, with no markdown code blocks formatting (e.g., do not wrap in ` ` ` json), just the raw JSON: \\
        \{ \\
        \ \ \ \ "essential\_property": "...", \\
        \ \ \ \ "solution\_principle": "...", \\
        \ \ \ \ "core\_concept": "..." \\
        \}
        \\
        \textbf{\#\# Problem:} \\
        {[}PROBLEM{]}
    \end{boxK}
    \vspace{-0.1in}
    \caption{Prompt for generating teacher mathematical primitives.}
    \label{fig:teacher_primitive_prompt}
\end{figure}

\begin{figure}[h]
    \centering
    \begin{boxK}[colback=royalblue!5!white, colframe=royalblue!75!black, fontupper=\ttfamily\tiny]
        \ You are a math problem solver. Below is a problem. Produce a concise high-level solution plan describing the major procedural steps needed to solve it.
        \\
        \textbf{\#\# HARD CONSTRAINTS:}
        \begin{itemize}[leftmargin=0.15in,nosep]
            \item Give 3--6 major steps in the order they should be carried out.
            \item Each step should state the intended mathematical operation, construction, theorem application, or reduction, without carrying out the corresponding derivation.
            \item Do NOT provide algebraic calculations, intermediate numerical results, detailed derivations, or the final answer.
            \item Keep the overall level of detail concise and comparable to a short conceptual guidance.
        \end{itemize}

        \textbf{\#\# Problem:} \\
        {[}PROBLEM{]}
    \end{boxK}
    \vspace{-0.1in}
    \caption{Prompt for generating teacher solution plans from language-only mathematics problems.}
    \label{fig:teacher_plan_prompt}
\end{figure}

\clearpage

\subsection{Additional Experiment Results}
\label{app:add-diagnosis-exp}
\subsubsection{Performance across Primitive Families}
To provide a more fine-grained view of model behavior, we further break down performance by primitive family. Table~\ref{tab:gap-family} reports results on \hlei{} separately for \textbf{Recast}, \textbf{Witness}, and \textbf{Argument} primitives across all four dimensions of mathematical understanding. This breakdown complements the aggregate results in Table~\ref{tab:digest-eval} in the main text by showing how the observed capability gaps vary across different forms of structural reasoning.

\begin{table*}[htbp]
\centering
\footnotesize
\setlength{\tabcolsep}{3.5pt}
\renewcommand{\arraystretch}{1.15}

\begin{tabular}{
@{}l
*{3}{S[table-format=3.2]}
@{\hspace{0.55em}}
*{3}{S[table-format=3.2]}
@{\hspace{0.55em}}
*{3}{S[table-format=3.2]}
@{\hspace{0.55em}}
*{3}{S[table-format=3.2]}
@{}
}
\toprule
& \multicolumn{3}{c}{\textbf{Discovery}}
& \multicolumn{3}{c}{\textbf{Generation}}
& \multicolumn{3}{c}{\textbf{Digestion}}
& \multicolumn{3}{c}{\textbf{Execution}} \\
\cmidrule(lr){2-4}
\cmidrule(lr){5-7}
\cmidrule(lr){8-10}
\cmidrule(lr){11-13}

\textbf{Model}
& {\textbf{R}} & {\textbf{W}} & {\textbf{A}}
& {\textbf{R}} & {\textbf{W}} & {\textbf{A}}
& {\textbf{R}} & {\textbf{W}} & {\textbf{A}}
& {\textbf{R}} & {\textbf{W}} & {\textbf{A}} \\
\midrule

\rowcolor{gray!8}[0pt][0pt]
\multicolumn{13}{@{}l}{\textit{OpenAI}} \\

gpt-5.4
& \num{84.97} & \num{72.73} & \num{71.43}
& \num{68.63} & \num{54.55} & \num{71.43}
& \num{100.00} & \num{100.00} & \num{100.00}
& \num{86.27} & \num{95.45} & \num{71.43} \\

gpt-5.4-mini
& \num{61.44} & \num{54.55} & \num{28.57}
& \num{51.63} & \num{40.91} & \num{57.14}
& \num{97.39} & \num{100.00} & \num{100.00}
& \num{70.59} & \num{81.82} & \num{71.43} \\

gpt-5.4-nano
& \num{43.79} & \num{27.27} & \num{42.86}
& \num{46.41} & \num{31.82} & \num{42.86}
& \num{96.73} & \num{100.00} & \num{100.00}
& \num{69.28} & \num{72.73} & \num{42.86} \\

gpt-oss-20b
& \num{36.60} & \num{13.64} & \num{57.14}
& \num{42.48} & \num{40.91} & \num{71.43}
& \num{91.50} & \num{90.91} & \num{85.71}
& \num{60.99} & \num{72.73} & \num{28.57} \\

\addlinespace[2pt]

\rowcolor{gray!8}[0pt][0pt]
\multicolumn{13}{@{}l}{\textit{Qwen}} \\

Qwen3.6-27B
& \num{28.76} & \num{0.00} & \num{14.29}
& \num{53.59} & \num{45.45} & \num{57.14}
& \num{90.85} & \num{100.00} & \num{100.00}
& \num{78.43} & \num{77.27} & \num{85.71} \\

Qwen3.5-27B
& \num{29.41} & \num{18.18} & \num{42.86}
& \num{47.71} & \num{40.91} & \num{71.43}
& \num{92.81} & \num{95.45} & \num{100.00}
& \num{70.59} & \num{81.82} & \num{71.43} \\

Qwen3.5-9B
& \num{14.38} & \num{9.09} & \num{0.00}
& \num{38.56} & \num{27.27} & \num{71.43}
& \num{78.43} & \num{86.36} & \num{85.71}
& \num{59.48} & \num{72.73} & \num{71.43} \\

Qwen3.5-4B
& \num{7.19} & \num{0.00} & \num{14.29}
& \num{26.14} & \num{27.27} & \num{28.57}
& \num{69.93} & \num{72.73} & \num{85.71}
& \num{54.25} & \num{68.18} & \num{57.14} \\

\addlinespace[2pt]

\rowcolor{gray!8}[0pt][0pt]
\multicolumn{13}{@{}l}{\textit{DeepSeek-R1 Distill}} \\

R1-0528-8B
& \num{5.88} & \num{4.55} & \num{28.57}
& \num{15.69} & \num{27.27} & \num{14.29}
& \num{67.97} & \num{81.82} & \num{42.86}
& \num{37.25} & \num{54.55} & \num{42.86} \\

R1-Distill-32B
& \num{6.54} & \num{4.55} & \num{0.00}
& \num{15.03} & \num{31.82} & \num{42.86}
& \num{64.05} & \num{68.18} & \num{85.71}
& \num{41.83} & \num{59.09} & \num{28.57} \\

R1-Distill-14B
& \num{4.58} & \num{4.55} & \num{0.00}
& \num{15.69} & \num{18.18} & \num{14.29}
& \num{68.63} & \num{54.55} & \num{71.43}
& \num{37.25} & \num{54.55} & \num{28.57} \\

R1-Distill-7B
& \num{7.84} & \num{4.55} & \num{0.00}
& \num{13.73} & \num{9.09} & \num{14.29}
& \num{49.02} & \num{50.00} & \num{42.86}
& \num{30.07} & \num{45.45} & \num{42.86} \\

\bottomrule
\end{tabular}

\caption{Performance on \hlei{} across the four dimensions of mathematical understanding, stratified by primitive family: \textbf{R} = Recast, \textbf{W} = Witness, and \textbf{A} = Argument.}
\label{tab:gap-family}
\end{table*}

\subsubsection{Reasoning Length and Generation Behavior}
\begin{figure}[htbp]
\vspace{-0.15in}
  \centering
  \begin{subfigure}[t]{0.35\textwidth}
    \includegraphics[width=\linewidth]{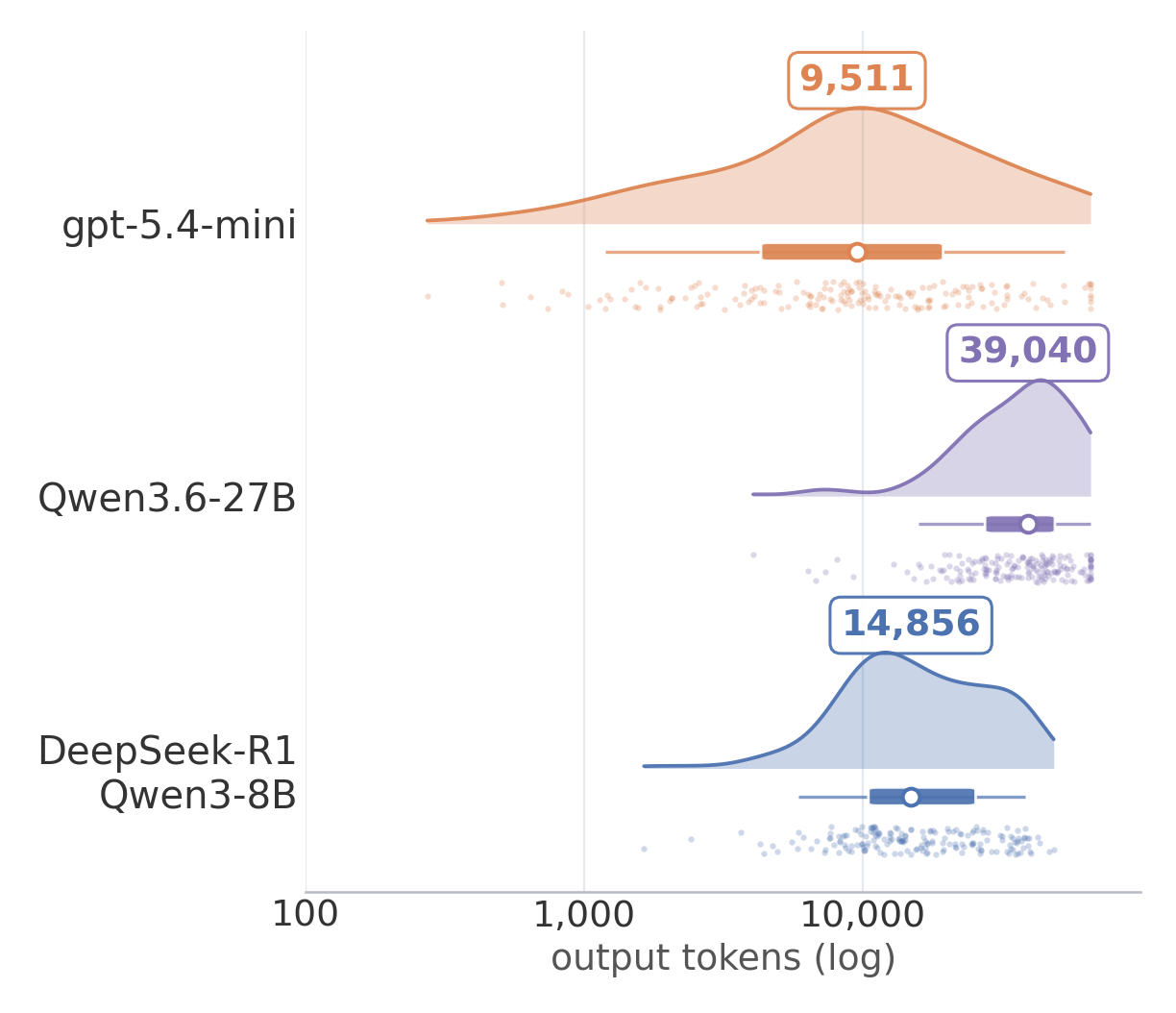}
    \caption{All rollouts}
    \label{fig:tok-dc}
  \end{subfigure}
  \begin{subfigure}[t]{0.35\textwidth}
    \includegraphics[width=\linewidth]{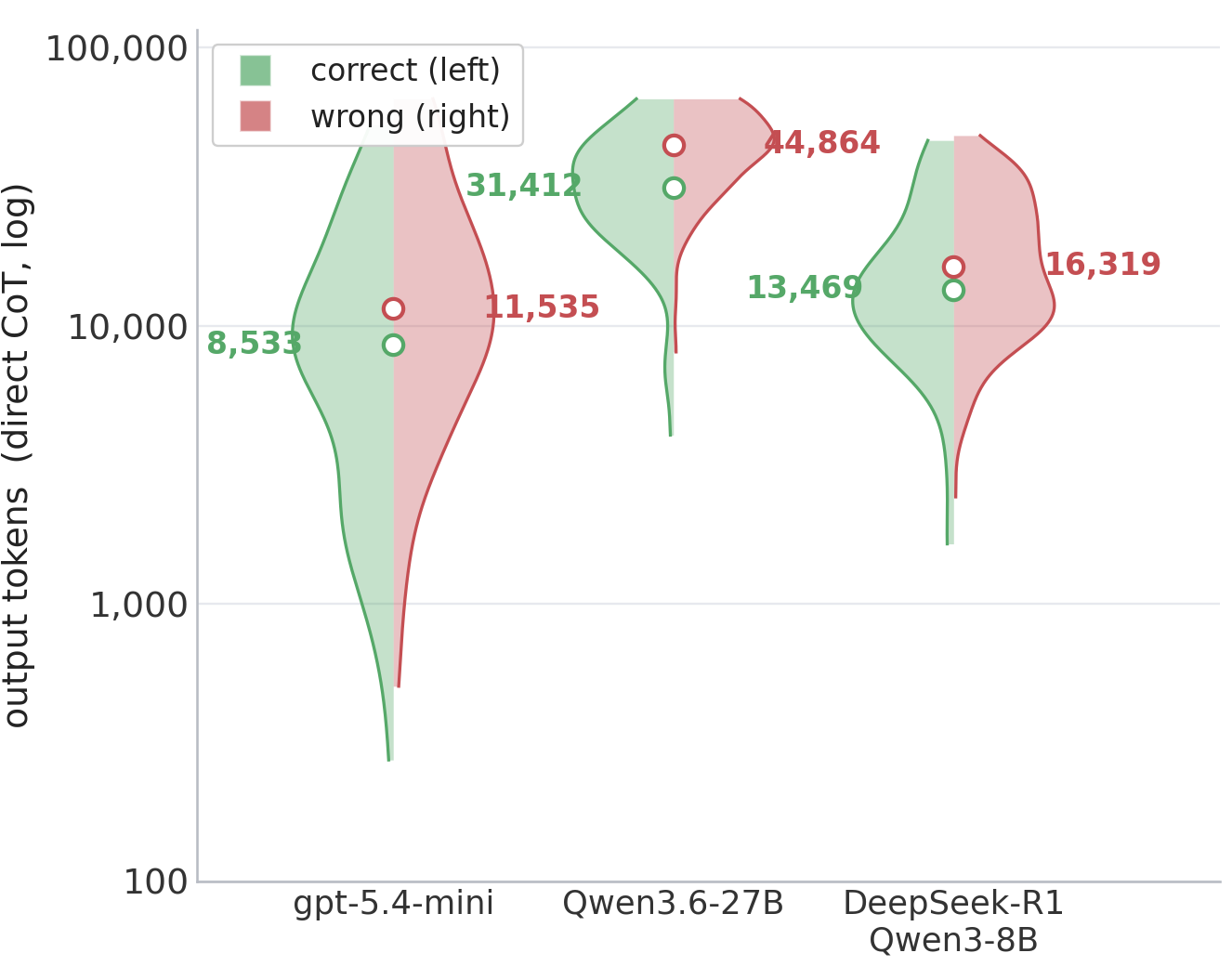}
    \caption{By correctness}
    \label{fig:tok-correct}
  \end{subfigure}
  \vspace{-0.12in}
  \caption{Output-token distributions of representative models from each model family for \textbf{Generation}. Left: all rollouts. Right: rollouts separated by final-answer correctness.}
  \label{fig:token-distributions}
  \vspace{-0.1in}
\end{figure}

Figure~\ref{fig:token-distributions} compares the output-token distributions of representative models during Generation. The left panel shows the overall distribution across all rollouts, while the right panel separates rollouts by final-answer correctness, providing a view of how reasoning length varies across models and between successful and unsuccessful solutions.

\subsubsection{Execution under Different Primitive Sources}

Table~\ref{tab:execution-primitive-app} provides the full results underlying the Execution analysis in \S\ref{sec:dignos}, extending the comparison from the representative models in Table~\ref{tab:execution-primitive} to all 12 evaluated models. The same pattern holds consistently: conditioning on self-generated primitives provides little or no benefit and often degrades performance, whereas externally supplied teacher primitives substantially improve Execution across model families, with gold primitives yielding the largest gains. These results further support that the rescue effect is not a consequence of simply decomposing generation into a discover-then-execute pipeline; rather, successful execution depends on supplying the relevant mathematical structure that the model often fails to identify on its own.

\sisetup{
    detect-family = true,
    detect-weight = true
}

\begin{table}[htbp]
\centering
\footnotesize
\setlength{\tabcolsep}{3pt}
\renewcommand{\arraystretch}{1.15}

\begin{tabular}{@{}lcccc@{\hspace{4pt}}}
\toprule
\multirow{2}{*}{\textbf{Model}}
& \multirow{2}{*}{\textbf{Generation}}
& \multicolumn{3}{c}{\textbf{Execution}} \\
\cmidrule(lr){3-5}

&
& {\textbf{Self-Generated Primitive}}
& {\textbf{Teacher Primitive}}
& {\textbf{Gold Primitive}} \\
\midrule

\rowcolor{gray!8}[0pt][4pt]
\multicolumn{5}{@{}l}{\textit{OpenAI}} \\

gpt-5.4
& \num{67.03}
& \num{74.18}\chg{+7.15}
& ---$^\dagger$
& \num{86.81}\chg{+19.78} \\

gpt-5.4-mini
& \num{50.55}
& \num{48.90}\chg{-1.65}
& \num{63.74}\chg{+13.19}
& \num{71.98}\chg{+21.43} \\

gpt-5.4-nano
& \num{44.51}
& \num{46.15}\chg{+1.64}
& \num{60.99}\chg{+16.48}
& \num{68.68}\chg{+24.17} \\

gpt-oss-20b
& \num{43.41}
& \num{38.46}\chg{-4.95}
& \num{58.24}\chg{+14.83}
& \num{60.99}\chg{+17.58} \\

\addlinespace[2pt]

\rowcolor{gray!8}[0pt][4pt]
\multicolumn{5}{@{}l}{\textit{Qwen}} \\

Qwen3.6-27B
& \num{52.75}
& \num{41.21}\chg{-11.54}
& \num{64.84}\chg{+12.09}
& \num{78.57}\chg{+25.82} \\

Qwen3.5-27B
& \num{47.80}
& \num{36.81}\chg{-10.99}
& \num{58.24}\chg{+10.44}
& \num{71.98}\chg{+24.18} \\

Qwen3.5-9B
& \num{38.46}
& \num{23.08}\chg{-15.38}
& \num{52.75}\chg{+14.29}
& \num{61.54}\chg{+23.08} \\

Qwen3.5-4B
& \num{26.37}
& \num{17.03}\chg{-9.34}
& \num{42.31}\chg{+15.94}
& \num{56.04}\chg{+29.67} \\

\addlinespace[2pt]

\rowcolor{gray!8}[0pt][4pt]
\multicolumn{5}{@{}l}{\textit{DeepSeek-R1 Distill}} \\

R1-0528-Qwen3-8B
& \num{17.03}
& \num{10.99}\chg{-6.04}
& \num{31.32}\chg{+14.29}
& \num{39.56}\chg{+22.53} \\

R1-Distill-Qwen-32B
& \num{18.13}
& \num{17.58}\chg{-0.55}
& \num{37.91}\chg{+19.78}
& \num{43.41}\chg{+25.28} \\

R1-Distill-Qwen-14B
& \num{15.93}
& \num{14.84}\chg{-1.09}
& \num{31.87}\chg{+15.94}
& \num{39.01}\chg{+23.08} \\

R1-Distill-Qwen-7B
& \num{13.19}
& \num{10.44}\chg{-2.75}
& \num{19.78}\chg{+6.59}
& \num{32.42}\chg{+19.23} \\

\bottomrule
\end{tabular}

\caption{
Generation and Execution performance on \hlei{} under different settings. Execution is evaluated with self-generated, teacher-provided, and gold primitives. Percentage-point changes are reported relative to Generation. $^\dagger$The teacher-primitive result for gpt-5.4 is omitted because gpt-5.4 serves as the teacher model.
}
\label{tab:execution-primitive-app}
\end{table}

\sisetup{
    detect-family = true,
    detect-weight = true
}

\begin{table}[htbp]
\centering
\footnotesize
\setlength{\tabcolsep}{4.5pt}
\renewcommand{\arraystretch}{1.15}

\begin{tabular}{@{}lccccc@{\hspace{4pt}}}
\toprule
\multirow{2}{*}{\textbf{Model}}
& \multirow{2}{*}{$\boldsymbol{n_{\mathrm{fail}}}$}
& \multicolumn{4}{c}{\textbf{Composition of Generation Failures (\%)}} \\
\cmidrule(lr){3-6}

&
& \textbf{$D^-E^+$}
& \textbf{$D^-E^-$}
& \textbf{$D^+E^-$}
& \textbf{$D^+E^+$} \\[-1pt]

&
& {\scriptsize Discovery-limited}
& {\scriptsize Capability-limited}
& {\scriptsize Execution-limited}
& {\scriptsize Utilization-limited} \\
\midrule

\rowcolor{gray!8}[0pt][4pt]
\multicolumn{6}{@{}l}{\textit{OpenAI}} \\

gpt-5.4
& \num{60}
& \num{15.0}
& \num{10.0}
& \num{21.7}
& \num{53.3} \\

gpt-5.4-mini
& \num{90}
& \num{28.9}
& \num{21.1}
& \num{26.7}
& \num{23.3} \\

gpt-5.4-nano
& \num{101}
& \num{38.6}
& \num{31.7}
& \num{18.8}
& \num{10.9} \\

gpt-oss-20b
& \num{103}
& \num{35.0}
& \num{45.6}
& \num{9.7}
& \num{9.7} \\

\addlinespace[2pt]

\rowcolor{gray!8}[0pt][4pt]
\multicolumn{6}{@{}l}{\textit{Qwen}} \\

Qwen3.6-27B
& \num{86}
& \num{48.8}
& \num{33.7}
& \num{7.0}
& \num{10.5} \\

Qwen3.5-27B
& \num{95}
& \num{42.1}
& \num{30.5}
& \num{15.8}
& \num{11.6} \\

Qwen3.5-9B
& \num{112}
& \num{44.6}
& \num{42.9}
& \num{6.2}
& \num{6.2} \\

Qwen3.5-4B
& \num{134}
& \num{42.5}
& \num{52.2}
& \num{3.7}
& \num{1.5} \\

\addlinespace[2pt]

\rowcolor{gray!8}[0pt][4pt]
\multicolumn{6}{@{}l}{\textit{DeepSeek-R1 Distill}} \\

R1-0528-Qwen3-8B
& \num{151}
& \num{28.5}
& \num{65.6}
& \num{2.6}
& \num{3.3} \\

R1-Distill-Qwen-32B
& \num{149}
& \num{33.6}
& \num{61.7}
& \num{2.0}
& \num{2.7} \\

R1-Distill-Qwen-14B
& \num{153}
& \num{30.7}
& \num{68.0}
& \num{1.3}
& \num{0.0} \\

R1-Distill-Qwen-7B
& \num{158}
& \num{26.6}
& \num{68.4}
& \num{3.2}
& \num{1.9} \\

\bottomrule
\end{tabular}

\caption{Composition of direct Generation failures on \hlei{}. Each failed problem is categorized by whether the model discovers the gold
primitive from the question ($D$; $\tau=0.8$) and whether it can execute the primitive when provided ($E$).}
\label{tab:failure-composition}
\end{table}
\subsubsection{Failure Decomposition and Rescue Analysis} 

Table~\ref{tab:failure-composition} provides the full model-wise breakdown underlying Figure~\ref{fig:failure-decomposition}. Across model families, Generation failures are concentrated in the $D^{-}$ regimes. This decomposition further clarifies the role of primitive discovery: even when overall solving ability is relatively strong, a substantial portion of failures arises because the model cannot independently identify the relevant mathematical structure, despite being able to execute it once supplied.

We further examine how Digestion relates to a model's ability to make use of externally supplied primitive guidance. Table~\ref{tab:rescue-gating} partitions initially failed problems according to whether the model can correctly recover the primitive from the reference solution, and then measures the rescue rate after supplying the teacher primitive. For most models, digestible failures are substantially more likely to be rescued than undigestible ones, suggesting that the ability to recover mathematical structure from a completed solution is closely related to the ability to incorporate that structure into downstream reasoning. This relationship is particularly clear for the Qwen3.5 models, where digestible problems exhibit rescue-rate advantages of 17.6--28.6 percentage points. The pattern is weaker for the DeepSeek-R1 distill models and is noisy for models with very few undigestible examples, indicating that Digestion and successful execution are related but distinct capabilities.

\begin{table}[htbp]
\centering
\small
\setlength{\tabcolsep}{5pt}
\renewcommand{\arraystretch}{1.15}

\begin{tabular}{@{}lcccc@{\hspace{4pt}}}
\toprule
& \multirow{2}{*}{\textbf{Digestion}}
& \multicolumn{2}{c}{\textbf{Rescue Rate}}
& \multirow{2}{*}{\textbf{Gap}} \\
\cmidrule(lr){3-4}

\textbf{Model}
&
& \textbf{Digestible}
& \textbf{Undigestible}
& \\
\midrule

\rowcolor{gray!8}[0pt][4pt]
\multicolumn{5}{@{}l}{\textit{OpenAI}} \\

gpt-5.4-mini$^\dagger$
& \num{97.8}
& \num{42.5}
& \num{33.3}
& \num{+9.2} \\

gpt-5.4-nano$^\dagger$
& \num{97.3}
& \num{40.8}
& \num{0.0}
& \num{+40.8} \\

gpt-oss-20b$^\dagger$
& \num{91.2}
& \num{40.4}
& \num{11.1}
& \num{+29.3} \\

\addlinespace[2pt]

\rowcolor{gray!8}[0pt][4pt]
\multicolumn{5}{@{}l}{\textit{Qwen}} \\

Qwen3.6-27B$^\dagger$
& \num{92.3}
& \num{36.4}
& \num{66.7}
& \num{-30.3} \\

Qwen3.5-27B
& \num{93.4}
& \num{37.6}
& \num{10.0}
& \num{+27.6} \\

Qwen3.5-9B
& \num{79.7}
& \num{41.6}
& \num{13.0}
& \num{+28.6} \\

Qwen3.5-4B
& \num{70.9}
& \num{35.1}
& \num{17.5}
& \num{+17.6} \\

\addlinespace[2pt]

\rowcolor{gray!8}[0pt][4pt]
\multicolumn{5}{@{}l}{\textit{DeepSeek-R1 Distill}} \\

R1-0528-8B
& \num{68.7}
& \num{23.3}
& \num{16.7}
& \num{+6.6} \\

R1-Distill-32B
& \num{65.4}
& \num{32.6}
& \num{26.3}
& \num{+6.3} \\

R1-Distill-14B
& \num{67.0}
& \num{24.8}
& \num{23.1}
& \num{+1.7} \\

R1-Distill-7B
& \num{48.9}
& \num{19.7}
& \num{13.4}
& \num{+6.3} \\

\bottomrule
\end{tabular}

\caption{
Relationship between Digestion and rescue under teacher-primitive guidance.
Problems are partitioned by whether the model correctly recovers the primitive
from the reference solution (\textit{digestible}) or not (\textit{undigestible});
rescue rate is the fraction of initially incorrect problems solved after providing
the teacher primitive. Gap denotes the digestible--undigestible difference in rescue rate.
$^\dagger$Undigestible estimates are based on fewer than 20 problems and are therefore
high variance.
}
\label{tab:rescue-gating}
\end{table}

\subsubsection{Repairability across Failure Modes}
\label{app:repairability-failure-modes}

To complement the main-text analysis of discovery-limited and capability-limited failures, Table~\ref{tab:repair-quadrants-app} reports repair rates across all four initial failure quadrants of the base model. The distinction between $D^-E^+$ and $D^-E^-$ remains particularly consistent: across SFT, OPSD, and \abs{}, discovery-limited failures are repaired substantially more often than capability-limited failures. The remaining quadrants exhibit different repair profiles, with utilization-limited ($D^+E^+$) cases showing relatively high repair rates and execution-limited ($D^+E^-$) cases varying more strongly across post-training methods. These results provide a fuller view of how the initial capability decomposition relates to subsequent repairability, while reinforcing the main-text observation that failures with latent execution capacity are especially amenable to post-training.

\begin{table}[htbp]
\centering
\small
\setlength{\tabcolsep}{7pt}
\begin{tabular}{lrrr}
\toprule
\textbf{Base failure type} & \textbf{SFT} & \textbf{OPSD} & \textbf{\abs{}} \\
\midrule
$D^-E^+$: discovery-limited
& \textbf{20.4\%}
& \textbf{21.1\%}
& \textbf{23.8\%} \\

$D^-E^-$: capability-limited
& 6.8\%
& 8.2\%
& 10.2\% \\

$D^+E^-$: execution-limited
& 11.1\%
& 0.0\%
& 18.5\% \\

$D^+E^+$: utilization-limited
& 35.0\%
& 30.0\%
& 45.0\% \\

\bottomrule
\end{tabular}
\caption{
Post-training repair rates across initial \hlei{} failure quadrants. 
}
\label{tab:repair-quadrants-app}
\end{table}

\section{Post-Training Dataset Details}
\label{app:dataset-details}

Table~\ref{tab:dataset_stats} presents the detailed composition of the 709 Mathematics Ph.D.\ qualifying examination problems used for our post-training experiments, categorized by source university and mathematical domain. The problems span 1991--2026 and are strictly parsed into \LaTeX{}. We select qualifying-examination problems because their difficulty and open-ended reasoning demands closely match those of \hlei{}, requiring substantial mathematical maturity and structural reasoning rather than short-form answer prediction. Accordingly, the dataset consists entirely of open-ended proof problems without algorithmically verifiable short-answer labels. Each problem is paired with a human-authored reference proof (median length: 118 words) and a one-sentence primitive extracted from that proof by GPT-5.4-High using the curation prompt of Figure~\ref{fig:primitive_curation_prompt}.

\begin{table}[htbp]
\centering
\small
\renewcommand{\arraystretch}{1.15}
\begin{tabular}{@{}ll@{}}
\toprule
\textbf{Source University (1991--2026)} & \textbf{Problem Count} \\
\midrule
University of Oregon & 332 \\
UW Madison & 131 \\
Harvard University & 126 \\
UC Berkeley & 120 \\
\midrule
\textbf{Mathematical Domain} & \textbf{Problem Count} \\
\midrule
Algebra & 318 \\
Analysis & 206 \\
Topology & 97 \\
Geometry & 82 \\
Number Theory & 5 \\
Logic & 1 \\
\bottomrule
\end{tabular}
\vspace{-0.05in}
\caption{Detailed composition of the training corpus.}
\label{tab:dataset_stats}
\end{table}

\section{Post-Training Setup}
All \abs{} and OPSD runs use the same training recipe and differ in the privileged context provided to the teacher: \abs{} uses the mathematical primitive, whereas OPSD uses the reference solution. Training uses the 709 qualifying-examination problems, each paired with a one-sentence primitive. For each problem, the student generates one on-policy rollout from the problem alone using temperature $1.0$, top-$p$ $0.95$, top-$k$ $20$, a maximum of 4096 completion tokens, and a total context length of 6144. The teacher uses the same base model with the adapter disabled and is conditioned on the corresponding privileged context.

For \abs{}, we optimize $\mathcal{L}_{\abs{}}$ with top-$K$ support $K=128$ and clamp threshold $\tau=0.06$, applied to all generated completion tokens, including truncated rollouts. We train LoRA adapters with rank $64$ and $\alpha=128$ on all attention and MLP projections, including the linear-attention projections in Qwen3.5. Optimization uses fused AdamW without weight decay, a learning rate of $5\times10^{-6}$ with linear scheduling and no warmup, gradient clipping at $0.1$, and an effective batch size of $32$ with per-device batch size $1$ and gradient accumulation. Training uses bf16, gradient checkpointing, and seed $42$. Each model is trained for one epoch, corresponding to $22$--$23$ optimizer steps. On-policy rollouts are generated with a co-located vLLM engine synchronized with the student after every optimizer step.

\section{Post-Training Ablation Studies}
\label{app:ablations}

To validate the design choices of \abs{}, we conduct ablation studies on the Qwen3.5-9B baseline using \hlei{} ($n=182$). We organize the analysis around three questions: what information should be provided to the teacher as privilege, how that privileged guidance should be transferred to the student, and whether the primitive should instead be learned as an explicit prediction target. For all comparisons, we report Generation and Discovery together with absolute changes relative to the base model.

\begin{table}[htbp]
\centering
\small
\setlength{\tabcolsep}{8pt}
\renewcommand{\arraystretch}{1.2}
\begin{tabular}{lll}
\toprule
\textbf{Variant} & \textbf{Generation} & \textbf{Discovery} \\
\midrule

Qwen3.5-9B (base)
& \num{38.46}
& \num{13.19} \\
\midrule
Solution privilege
& \num{33.52}\chg{-4.95}
& \num{10.99}\chg{-2.20} \\

\rowcolor{gray!8}
\textbf{Primitive privilege (\abs{})}
& {\bfseries\num{43.96}}\chg{+5.49}
& {\bfseries\num{14.29}}\chg{+1.10} \\

\bottomrule
\end{tabular}

\caption{
Ablation on the source of privileged information using Qwen3.5-9B. The training objective is held fixed while the teacher is conditioned on either the full reference solution or the mathematical primitive.
}
\label{tab:ablation-privilege}
\end{table}
\paragraph{Teacher Conditioning}
Table~\ref{tab:ablation-privilege} isolates the effect of the teacher's privileged input. In both variants, the student follows the same on-policy solution rollout and is trained with the same \abs{} objective; the only difference is whether the teacher conditions on the full reference solution or on the mathematical primitive. Under this controlled comparison, solution conditioning reduces Generation from \num{38.46} to \num{33.52}, whereas primitive conditioning improves it to \num{43.96}. Discovery follows the same trend. This shows that the gain does not arise from privileged conditioning alone: the primitive provides a more targeted supervisory signal than the full solution.

\paragraph{Transfer Mechanism}
Table~\ref{tab:ablation-override} fixes the mathematical primitive as the teacher's privileged input and keeps the student rollout in solution space, while varying how the primitive-conditioned teacher signal is transferred. The \emph{rKL, no cap} variant applies reverse KL without restricting teacher-induced changes, providing the unconstrained baseline. The \emph{fKL, capped} variant instead uses forward KL on the teacher's top-$K$ support while retaining the same cap, testing whether the benefit comes simply from limiting large contributions rather than from the asymmetric reverse-KL formulation. Both variants underperform the base model, with Generation dropping to \num{31.87} and \num{36.81}, respectively.

\begin{table}[h]
\centering
\small
\setlength{\tabcolsep}{8pt}
\renewcommand{\arraystretch}{1.2}
\begin{tabular}{lll}
\toprule
\textbf{Transfer Mechanism} & \textbf{Generation} & \textbf{Discovery} \\
\midrule

Qwen3.5-9B (base)
& \num{38.46}
& \num{13.19} \\
\midrule
rKL, no cap
& \num{31.87}\chg{-6.59}
& \num{10.44}\chg{-2.75} \\

fKL, capped
& \num{36.81}\chg{-1.65}
& \num{12.09}\chg{-1.10} \\

Divergence control
& \num{34.62}\chg{-3.85}
& \num{12.09}\chg{-1.10} \\

Entropy control
& \num{31.87}\chg{-6.59}
& \num{14.29}\chg{+1.10} \\

\rowcolor{gray!8}
\textbf{Bounded override (\abs{})}
& {\bfseries\num{43.96}}\chg{+5.49}
& {\bfseries\num{14.29}}\chg{+1.10} \\

\bottomrule
\end{tabular}

\caption{
Ablation on mechanisms for transferring primitive-guided teacher supervision using Qwen3.5-9B. All variants use the mathematical primitive as privileged information.
}
\label{tab:ablation-override}
\end{table}

\emph{Divergence control} tests whether harmful transfer can be identified purely by the magnitude of teacher--student disagreement. Let
\begin{align*}
    D_t = D_{\mathrm{KL}}\!\left( \bar{\pi}_S(\cdot \mid \hat{y}_{<t},x) \;\middle\|\; \bar{\pi}_T(\cdot \mid \hat{y}_{<t},x,p) \right),
\end{align*}
where both distributions are renormalized. Since the divergence distribution is highly concentrated in a small tail, we mask tokens whose divergence exceeds the offline $95$th-percentile threshold:
\begin{align*}
m_t^{\mathrm{div}} = \mathbf{1}\!\left[D_t \le \tau_D\right], \qquad \tau_D = 0.05.
\end{align*}
This removes only about $5\%$ of tokens but approximately $86\%$ of the total divergence mass, partially recovering Generation from \num{31.87} to \num{34.62}.

\emph{Entropy control} tests a finer hypothesis: whether harmful overrides and useful primitive guidance can be separated within this high-divergence tail using the student's confidence. We define the student's full-vocabulary entropy as
\begin{align*}
    H_t = -\sum_v \pi_S(v\mid \hat{y}_{<t},x) \log \pi_S(v\mid \hat{y}_{<t},x),
\end{align*}
and mask only high-divergence, low-entropy positions:
\begin{align*}
    m_t^{\mathrm{ent}} = \mathbf{1}\!\left[ D_t \le \tau_D \;\lor\; H_t \ge \tau_H \right], \qquad \tau_H = 0.86.
\end{align*}
This preserves high-divergence positions where the student is uncertain, under the hypothesis that these are more likely to contain useful primitive-specific guidance. However, despite masking fewer tokens, Generation returns to \num{31.87}, indicating that neither divergence magnitude nor student confidence cleanly separates beneficial from harmful teacher intervention.

In contrast, \abs{} avoids making a whole-token keep-or-drop decision. Its bounded override operates within each token distribution, retaining supervision at every position while limiting how strongly individual teacher preferences can suppress student-supported choices. This is the only transfer mechanism in the comparison that improves over the base model, reaching \num{43.96} Generation.

\paragraph{Rollout Target}
Table~\ref{tab:ablation-rollout} isolates whether the mathematical primitive should serve as privileged guidance or as the student's rollout target. In the \abs{} setting, the student continues to generate solutions while the primitive is available only to the teacher. We compare this against variants in which the student instead generates the primitive itself under uncapped reverse KL, capped forward KL, or the same capped reverse-KL objective used by \abs{}. Primitive-target rollouts remain weak across all three objectives. Even under the \abs{} objective, changing only the rollout target from solution to primitive yields \num{39.01} Generation, compared with \num{43.96} when the primitive instead conditions the teacher while the student generates solutions. This indicates that the primitive is more effective as privileged guidance that shapes solution reasoning than as the object the student is directly optimized to produce.

\begin{table}[htbp]
\centering
\small
\setlength{\tabcolsep}{7pt}
\renewcommand{\arraystretch}{1.2}
\begin{tabular}{lcc}
\toprule
\textbf{Rollout Target / Objective} & \textbf{Generation} & \textbf{Discovery} \\
\midrule

Qwen3.5-9B (base)
& \num{38.46}
& \num{13.19} \\

\rowcolor{gray!8}
\textbf{\abs{}}
& {\bfseries\num{43.96}}\chg{+5.49}
& {\bfseries\num{14.29}}\chg{+1.10} \\

\midrule

rKL, no cap
& \num{32.97}\chg{-5.49}
& \num{13.19}\chg{+0.00} \\

fKL, capped
& \num{33.52}\chg{-4.95}
& \num{9.34}\chg{-3.85} \\

rKL, capped
& \num{39.01}\chg{+0.55}
& \num{12.09}\chg{-1.10} \\

\bottomrule
\end{tabular}

\caption{
Ablation on the student rollout target using Qwen3.5-9B. The \abs{} baseline keeps the student in solution-generation space and uses the primitive only as privileged teacher information; the remaining variants instead train the student to generate the primitive itself under different distillation objectives.
}
\label{tab:ablation-rollout}
\end{table}

\paragraph{Alternative Training Objectives}
Table~\ref{tab:ablation-objective} compares \abs{} against alternative
post-training objectives on the same Qwen3.5-9B backbone. We separate the
comparison into teacher-guided distillation objectives and direct supervised
fine-tuning objectives.

Among the distillation baselines, \emph{TOP-D}~\citep{xie2026trust} is a
policy-gradient on-policy distillation method based on a bounded proximal
reward. We adapt it to our setting using the same 709-problem training corpus,
primitive privilege, and self-teacher as \abs{}, with $G=4$ student rollouts
per prompt, proximal reward
\[
\tilde{r}_t
=
\log\!\left(\alpha \rho_t + 1-\alpha\right),
\qquad
\alpha = 0.1,
\]
followed by token-level length-normalized future returns and group-normalized
advantages. Despite operating with the same privileged signal, TOP-D reduces
Generation to \num{31.87} and Discovery to \num{9.34}, indicating that the
primitive alone is insufficient without an appropriate transfer mechanism.

\emph{Dual-channel} is an alternative design motivated by our analysis of
rescued trajectories, which suggested that primitive guidance primarily acts
through early route selection and later adjudication. The route channel applies
reverse KL only to the first 512 completion tokens. The adjudication channel
operates at student-generated self-doubt forks (e.g., ``Wait'', ``But'',
``However'', ``Actually'') and asks the student to sample candidate
continuations. The primitive-conditioned teacher provides only a preference
over these student-generated candidates rather than generating tokens itself.
Given teacher and frozen-base scores $z_T$ and $z_0$, we form the
base-corrected teacher preference
\[
q_T
=
\operatorname{softmax}
\left(
\frac{z_T-z_0}{0.2}
\right),
\]
and optimize
\[
\mathcal{L}_{\mathrm{adj}}
=
D_{\mathrm{KL}}(q_T \parallel q_S),
\]
with weight $0.3$ in addition to the early-route reverse-KL term. Because this
variant uses a larger 1,692-problem pool, full-length primitive annotations,
and 8k-token completions, it should be interpreted as a reference alternative
rather than a strict single-variable ablation. It nevertheless substantially
underperforms both the base model and \abs{}.

We separately compare direct supervised objectives. \emph{Solution SFT}
fine-tunes the model on the human-authored reference solutions, while
\emph{primitive SFT} directly trains the model to generate the mathematical
primitive. Solution SFT lowers Generation to \num{32.42}, and primitive SFT
performs worst overall, reducing Generation and Discovery to \num{28.02} and
\num{2.75}, respectively. Together, these results show that neither alternative
teacher-guided distillation nor direct supervision on solutions or primitives
substitutes for the combination used by \abs{}: primitive-conditioned teacher
guidance transferred through bounded override while the student remains in
solution-generation space.

\begin{table}[htbp]
\centering
\small
\setlength{\tabcolsep}{7pt}
\renewcommand{\arraystretch}{1.2}
\begin{tabular}{lll}
\toprule
\textbf{Training Objective} & \textbf{Generation} & \textbf{Discovery} \\
\midrule

Qwen3.5-9B (base)
& \num{38.46}
& \num{13.19} \\

\rowcolor{gray!8}
\textbf{\abs{}}
& {\bfseries\num{43.96}}\chg{+5.49}
& {\bfseries\num{14.29}}\chg{+1.10} \\

\midrule
\multicolumn{3}{l}{\textit{Teacher-guided distillation}} \\

TOP-D~\citep{xie2026trust}
& \num{31.87}\chg{-6.59}
& \num{9.34}\chg{-3.85} \\

Dual-channel$^\dagger$
& \num{29.67}\chg{-8.79}
& \num{10.44}\chg{-2.75} \\

\midrule
\multicolumn{3}{l}{\textit{Direct supervised fine-tuning}} \\

Solution SFT
& \num{32.42}\chg{-6.04}
& \num{12.64}\chg{-0.55} \\

Primitive SFT
& \num{28.02}\chg{-10.44}
& \num{2.75}\chg{-10.44} \\

\bottomrule
\end{tabular}

\caption{
Comparison of alternative post-training objectives using Qwen3.5-9B.
Teacher-guided distillation baselines include TOP-D~\citep{xie2026trust} and
our dual-channel alternative, while the SFT baselines directly supervise either
the reference solution or the mathematical primitive. $^\dagger$Dual-channel
uses a larger 1,692-problem training pool, full-length primitive annotations,
and 8k-token completions, and is therefore included as a reference alternative
rather than a strict single-variable ablation.
}
\label{tab:ablation-objective}
\end{table}

\end{document}